\documentclass[10pt,twocolumn,letterpaper]{article}
\usepackage[accsupp]{axessibility}
\usepackage[]{iccv}      % To produce the REVIEW version
\usepackage{amsmath, amsthm, amssymb}

\usepackage{mathrsfs}
\usepackage{bm}
\usepackage{tcolorbox}
\usepackage{colortbl}

\usepackage{multirow}

\PassOptionsToPackage{table,x11names}{xcolor}
\usepackage{xcolor}

\usepackage{algorithm}
\usepackage{algorithmic}

\definecolor{iccvblue}{rgb}{0.21,0.49,0.74}
\usepackage[pagebackref,breaklinks,colorlinks,allcolors=iccvblue]{hyperref}

\def\paperID{4576} % *** Enter the Paper ID here
\def\confName{ICCV}
\def\confYear{2025}

\title{FG-OrIU: Towards Better Forgetting via Feature-Gradient Orthogonality for Incremental Unlearning}

\author{ \textbf{Qian Feng}$^{1}$, \textbf{JiaHang Tu}$^1$, \textbf{Mintong Kang}$^2$, \textbf{Hanbin Zhao}$^1$\thanks{Corresponding author: Hanbin Zhao $<$zhaohanbin@zju.edu.cn$>$.}, \textbf{Chao Zhang}$^{1}$, \textbf{Hui Qian}$^{1}$ \\
 $^1$College of Computer Science and Technology, Zhejiang University ~~ \\ $^2$Department of Computer Science University of Illinois at Urbana-Champaign
}

\begin{document}
% \maketitle
% \twocolumn[{
% 	\renewcommand\twocolumn[1][]{#1}
        \maketitle
% 	\centering
% 	\vspace{-0.7cm}
% 	\includegraphics[width=1\textwidth]{fig/11.pdf} 
% 	\vspace{-3.0cm}
% 	\captionof{figure}{The reconstructed images using Deep Image Prior (DIP) \cite{ulyanov2018deep} with the last block's features as the training target. Columns $1$ is the original image from $D_{f}$, and Columns $2$ shows the results of feature restoration from the pre-trained model before forgetting, which represents the upper bound for feature restoration. Columns $3$-$7$ represent the results after applying different unlearning methods to the pre-trained model, where it can be observed that these methods only blur the features, a phenomenon we refer to as \textit{\textbf{superficial forgetting}}. In contrast, our method, shown in the last column, thoroughly eliminates the information about the original image, resulting in the restoration of features that generate an image completely composed of noise, which we refer to as \textit{\textbf{real forgetting}}.}
% 	\vspace{+0.9cm}
% 	\label{fig:figure1}
% }]

\begin{abstract}
Incremental unlearning (IU) is critical for pre-trained models to comply with sequential data deletion requests, yet existing methods primarily suppress parameters or confuse knowledge without explicit constraints on both feature and gradient level, resulting in \textit{superficial forgetting} where residual information remains recoverable. This incomplete forgetting risks security breaches and disrupts retention balance, especially in IU scenarios. We propose FG-OrIU (\textbf{F}eature-\textbf{G}radient \textbf{Or}thogonality for \textbf{I}ncremental \textbf{U}nlearning), the first framework unifying orthogonal constraints on both features and gradients level to achieve deep forgetting, where the forgetting effect is irreversible. FG-OrIU decomposes feature spaces via Singular Value Decomposition (SVD), separating forgetting and remaining class features into distinct subspaces. It then enforces dual constraints: feature orthogonal projection on both forgetting and remaining classes, while gradient orthogonal projection prevents the reintroduction of forgotten knowledge and disruption to remaining classes during updates. Additionally, dynamic subspace adaptation merges newly forgetting subspaces and contracts remaining subspaces, ensuring a stable balance between removal and retention across sequential unlearning tasks. Extensive experiments demonstrate the effectiveness of our method.
\end{abstract}    
\section{Introduction}
\label{sec:intro}

In recent years, \textit{pre-training with fine-tuning} \cite{zhang2025parameter,han2024parameter,xu2023parameter} has dominated model development. However, large-scale datasets often contain harmful or private content, necessitating mechanisms to erase specific data traces. Machine Unlearning (MU) \cite{qin2024machine,cao2015towards,guo2019certified,shi2025redefining,bonato2024retain,foster2024fast,chundawat2023can,huang2024unified,bourtoule2021machine,shaik2024exploring,xu2024machine} addresses this by enabling pre-trained models (PTMs) to comply with deletion mandates like GDPR (General data protection regulation)’s “right to be forgotten” \cite{rosen2011right,regulation2018general}. While existing MU methods focus on static unlearning, forgetting a single or small set of classes in one task, real-world scenarios require Incremental Unlearning (IU) \cite{qureshi2025exploring,cheng2024remaining,zhao2025practical,zhao2024continual,wang2024has,gao2024practical,liu2022continual,chatterjee2024unified,kumaraveluuncle,chen2023unlearn,jang2022knowledge,grimes2024gone}. Regulatory shifts (e.g., evolving OpenAI safety policies\footnote{https://openai.com/safety/}, EU AI Act\footnote{https://artificialintelligenceact.eu/the-act/}) and dynamic content restrictions necessitate sequential unlearning of multiple classes/concepts without costly retraining. IU need not only erase targets in a single step but also preserve retained performance and ensure irreversible forgetting across tasks, making it critical for sustainable model governance in evolving ethical and legal landscapes.

\begin{figure}[t]
  \centering
   \includegraphics[width=1.0\columnwidth]{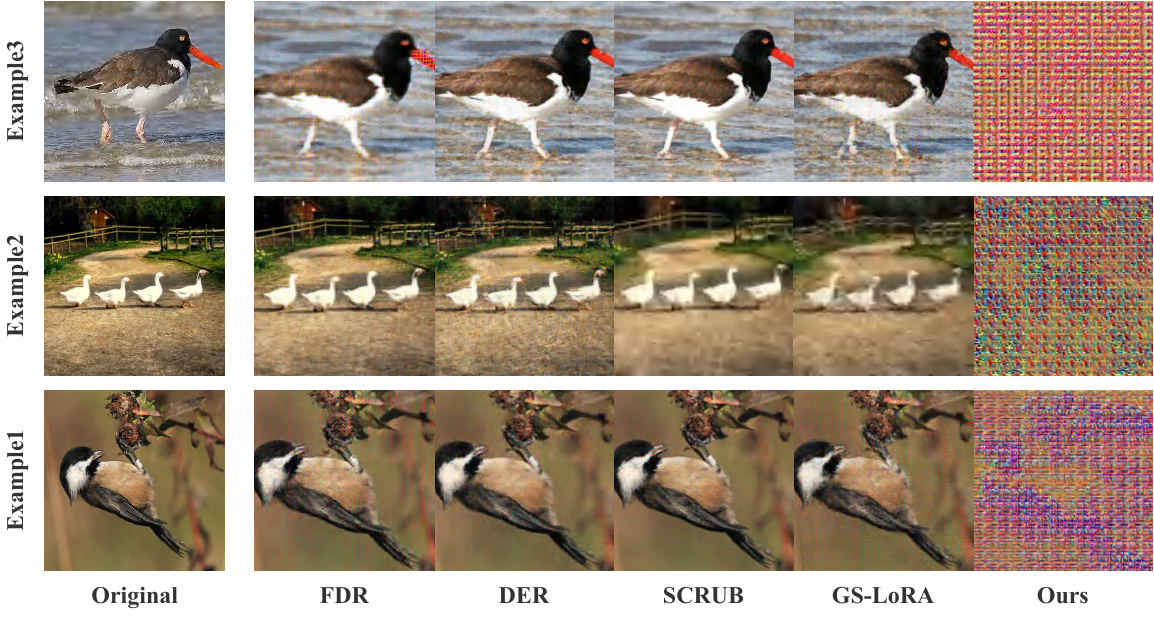}
   \vspace{-8mm}
   \caption{Here, we present three examples from forgetting classes, with the reconstructed images generated using Deep Image Prior (DIP) \cite{ulyanov2018deep}, where the last block's features from the unlearned model serve as the training target.}
   \label{fig:figure1}
   \vspace{-7mm}  
\end{figure}

IU methods broadly address two scenarios: (1) concept erasure in generative models, where sensitive concepts (e.g., nudity, violence) are iteratively removed from diffusion models \cite{tu2025sdwv,chin2023prompting4debugging}, and (2) instant- or class-level unlearning in vision classifiers, where PTMs forget specific classes (e.g., face identity, medical data) while retaining unrelated ones \cite{qin2024machine,jia2023model}. In this work, we focus on the latter, specifically targeting class-level IU in vision PTMs. Our goal is threefold: (a) to ensure that the model completely forgets the specified classes, (b) to robustly preserve the performance of remaining classes, and (c) to maintain the effectiveness of previous forgetting across sequential unlearning procedures.

Existing methods largely fall into two categories: \textit{parameter suppression}, which selectively weakens target class-related parameters \cite{foster2024fast,yan2022arcane,fan2023salun}, and \textit{knowledge confusion}, which distorts predictions via noisy labels or bad teacher distillation \cite{chundawat2023can,kurmanji2023towards,graves2021amnesiac}. However, our empirical findings reveal that these methods lead to \textit{superficial forgetting}, as the unlearned model still retains semantic information related to the forgetting classes, making the erased knowledge partially residual and recoverable. Specifically, reconstructing features from the last layer of the unlearned model using Deep Image Prior (DIP) \cite{ulyanov2018deep} produces blurred yet recognizable images (as illustrated in Figure \ref{fig:figure1}). Furthermore, fixing the backbone of the unlearned model and retraining only the classification head \cite{zhao2025practical,cheng2024remaining}, which recovers a portion of the original accuracy (as illustrated in Figure \ref{fig:recover_face_figure}). The blurred image and the partial recovery of performance both indicate that \textbf{existing methods merely degrade the features of forgetting classes rather than completely eliminating them.} The degraded features still contain residual class-specific information, which leads to potential recovery of the forgetting effect. Besides, these degraded features not only enable recovery but also risk entanglement with the remaining class features \cite{qin2024machine}. Such entanglement destabilizes decision boundaries, degrading performance on the remaining classes, particularly in IU scenarios where sequential deletion requests amplify error propagation \cite{wang2024has,gao2024practical}. To address these limitations and achieve \textit{deep forgetting}, we argue that unlearning should explicitly constrain both feature representations and gradient updates. Since model updates rely on both forward computations (feature extraction) and backward optimization (gradient updates), it is critical to simultaneously suppress the features of forgetting classes and prevent their reintroduction during training updates. In Table \ref{tab:tab1}, we summarize the differences between our methods and prior works.

To enforce explicit constraints on both feature representations and gradient updates, we draw inspiration from orthogonal mechanisms in related domains. For feature-level constraints, recent work in text-to-image models \cite{tu2025sdwv,yoon2024safree} demonstrates that orthogonalizing sensitive token embeddings effectively removes targeted concepts from diffusion outputs, suggesting geometric isolation as a viable strategy for feature suppression. For gradient-level constraints, inspired by \cite{yu2020gradient}, which reduces the mutual influence of different tasks in multi-task learning, and \cite{zhao2023rethinking,saha2021gradient}, which mitigates the forgetting of old tasks in continual learning, we leverage the insight that updating in directions orthogonal to old/other task subspaces helps alleviate catastrophic forgetting. While these orthogonal principles show promise in adjacent fields like generative model editing and multi-task/continual learning, their applicability to MU, particularly IU in vision PTMs, remains underexplored.

In this paper, we propose FG-OrIU (\textbf{F}eature-\textbf{G}radient \textbf{Or}thogonality for \textbf{I}ncremental \textbf{U}nlearning), a unified framework for IU in vision PTMs that enforces explicit constraints at both feature and gradient levels through three stages: \textbf{feature subspace decomposition, dual orthogonal projection, and dynamic subspace adaptation}. In \textbf{feature subspace decomposition}, we decompose the feature space of each layer into two subspaces via Singular Value Decomposition (SVD): the forgetting subspace $\mathcal{S}_{f}$, which captures the discriminative features of the classes to be removed, and the remaining subspace $\mathcal{S}_{r}$, which encodes the features of the remaining classes. In \textbf{dual orthogonal projection}, we freeze the backbone of PTM and introduce lightweight LoRA modules for efficient updates. Then, features of samples from forgetting classes are orthogonalized against $\mathcal{S}_{f}$ to eliminate correlations, while remaining classes features are aligned with $\mathcal{S}_{r}$ to minimize interference through minimizing the projection residual. Before backward optimization, gradients are first projected onto $\mathcal{S}_{f}$ to maximally perturb parameters critical for forgetting classes, then modified along the orthogonal complement of $\mathcal{S}_{r}$ to prevent contamination of retained features. In \textbf{dynamic subspace adaptation}, FOrIU dynamically expands $\mathcal{S}_{f}$ by merging new forgetting class subspaces and contracts $\mathcal{S}_{r}$ by removing overlapping features, ensuring cumulative forgetting without degrading retained performance. 

\begin{table}[t]
\centering
\caption{\textit{Position, Complete, and Recover} refers to, where the forgetting occurs, whether the forgetting is thorough, and whether the forgetting effect can be restored, respectively.}
\vspace{-2mm}
\label{tab:tab1}
\resizebox{\columnwidth}{!}{ % Resize the table to fit the text width
\small
\begin{tabular}{p{2cm}cccc}
\hline
 & \textit{Position} & \textit{Complete} & \textit{Recover} \\
\hline
\multicolumn{2}{l}{\textcolor{orange}{\textit{superficial forgetting}}} \\

\multirow{1}{*}{Prior works} & \multirow{1}{*}{Only Final Output}  & \multirow{1}{*}{No} & \multirow{1}{*}{Partial} \\
% \hline
% & Output & Degraded & No & Partial \\
\hline
\multicolumn{2}{l}{\textcolor{orange}{\textit{deep forgetting}}} \\

\multirow{1}{*}{Ours} & \multirow{1}{*}{Feature\&Gradient} & \multirow{1}{*}{Yes} & \multirow{1}{*}{No} \\
\hline
\end{tabular}
}
\vspace{-6mm}
\end{table}

\textbf{Our main contributions are threefold:}
\setlength{\leftmargini}{8pt} 

\begin{itemize} % 
% \vspace{-4mm} 
\item We reveal that existing IU methods achieve \textit{superficial forgetting} rather than \textit{deep forgetting}, as their partial feature suppression leaves residual information recoverable.
\item We propose the first framework unifying forgetting on both feature and gradient level via dual orthogonal projection, ensuring irreversible forgetting while preserving remaining class robustness.
\item Extensive experiments demonstrate superior performance in MU and IU scenarios across multiple benchmarks.
\end{itemize}

\vspace{-3mm}
\section{Related Work}
\vspace{-2mm}
\subsection{Machine Unlearning}
% \vspace{-2mm}
Machine unlearning (MU) aims to remove the influence of specific data from pre-trained models (PTMs) without retraining from scratch \cite{xu2023machine,zhang2023review}. Existing MU methods are broadly categorized into exact and approximate unlearning. Exact MU \cite{bourtoule2021machine,liu2021federaser} ensures statistical indistinguishability from models retrained on remaining data but is computationally prohibitive for large PTMs. Approximate MU relaxes this guarantee and is more practical. Current methods fall into two paradigms: (1) concept erasure in generative models \cite{tu2025sdwv,chin2023prompting4debugging}; and (2) class-/sample-wise unlearning in classification models, which modifies pre-trained weights to forget target classes while preserving others \cite{qin2024machine,jia2023model}. However, most studies focus on static unlearning, assuming a single-step forgetting task. Real-world applications, such as compliance with iterative AI safety policies (e.g., OpenAI’s content guidelines) or evolving regulations like the EU AI Act, require incremental unlearning (IU) to process sequential deletion requests efficiently. Directly applying static methods to IU often leads to cumulative errors, model collapse, and incomplete forgetting \cite{wang2024has,gao2024practical}, underscoring the need for dedicated frameworks to ensure compliance in dynamic regulatory environments.
\vspace{-2mm}
\subsection{Incremental Unlearning}
% \vspace{-2mm}
Existing MU methods, like \textit{parameter suppression} \cite{foster2024fast,yan2022arcane,fan2023salun} and \textit{knowledge confusion} methods \cite{chundawat2023can,kurmanji2023towards,graves2021amnesiac,gao2024practical,zhao2025practical,qureshi2025exploring,cheng2024remaining}, lead to \textit{superficial forgetting} due to the absence of direct constraints on the feature representation. In particular, the degraded features trap them in a dilemma between forgetting and remaining, which becomes even more severe when directly applied to IU. Meanwhile, some continual learning methods \cite{buzzega2020dark,kirkpatrick2017overcoming}, originally designed to retain old knowledge while dynamically learning new knowledge, also face inherent limitations in IU \cite{qureshi2025exploring}. Previous work focusing on IU has primarily relied on toy models, e.g., linear models \cite{ding2024fine}, or imposed heavy constraints, such as adding a class-specific synthetic signal in the pre-training stage \cite{shibata2021learning}, which is not feasible for PTMs. Moreover, while some focus on dynamically performing both continual learning and selective forgetting within learned knowledge \cite{kumaraveluuncle, liu2022continual,chatterjee2024unified}, our approach differs in that we emphasize achieving \textit{deep forgetting} when handling sequential unlearning tasks. That is, rather than alternating between learning and forgetting, the pre-trained model should completely forget unwanted knowledge while efficiently retaining the rest.

\vspace{-3mm}
\section{Problem Statement}
\label{sec:Preliminary}
\vspace{-1mm}
% We first introduce low-rank adaptation (LoRA) \cite{hu2022lora}, one of the most popular Parameter-Efficient Fine-Tuning (PEFT) methods used in our approach. Then, we provide the problem definition for both single-step forgetting and multi-step forgetting.

% \subsection{Low-Rank Adaption}
% \label{subsec:lora}
% When a pre-trained model (PTM) adapts to specific downstream tasks, \cite{hu2022lora} has demonstrated that weight updates in PTMs exhibit a low `intrinsic dimension'. For a pre-trained weight matrix $\mathbf{W}_{init} \in \mathbb{R}^{d \times k}$, LoRA constrains its update by representing it with a low-rank decomposition, $\mathbf{W_{init}}+\Delta \mathbf{W} = \mathbf{W_{init}} + \mathbf{A}\mathbf{B}$, where $\mathbf{A} \in \mathbb{R}^{d \times r}$, $\mathbf{B} \in \mathbb{R}^{r \times d}$, and the rank $r \ll \min\{d, k\}$. During training, $\mathbf{W_{init}}$ remains frozen while $\mathbf{A}$ and $\mathbf{B}$ are the only trainable parameters. Following prior works, we initialize $\mathbf{A}$ as $\mathbf{0}$ and initialize $\mathbf{B}$ with a Gaussian distribution, and then insert the LoRA modules into the FNN layers of each transformer block.

% \subsection{Problem Statement}
% \label{subsec:problem_statement}
% % \TODO{Combine together}

Following \cite{zhao2025practical,zhao2024continual}, we first formally define Machine Unlearning (MU), \ie, static unlearning scenario. We then extend this definition to Incremental Unlearning (IU), \ie, dynamic unlearning scenario, where deletion requests arrive sequentially.

Let $\mathcal{M}$ be a pre-trained model trained on the dataset $\mathbf{D}$. We denote the mapping relationship of the model as $f_\mathcal{M}(\mathcal{X}_\mathbf{D}) \rightarrow \mathcal{Y}_\mathbf{D}$, where $\mathcal{X}_\mathbf{D}$ and $\mathcal{Y}_\mathbf{D}$ represent the input set and output set, respectively. For MU, we partition $\mathbf{D}$ into $\mathbf{D_f}$ (forgetting classes) and $\mathbf{D_r}$ (remaining classes), under the constraint $|\mathbf{D_r}|+|\mathbf{D_f}|\ll |\mathbf{D}|$. The forgetting algorithm $\mathscr{F}$ produces $\mathcal{M}' = \mathscr{F}(\mathcal{M},\mathbf{D_f},\mathbf{D_r})$, altering the following mappings: 
\begin{equation}
f_{\mathcal{M}'}(\mathcal{X}_{\mathbf{D_f}}) \not \stackrel{}{\longrightarrow} \mathcal{Y}_{\mathbf{D_f}}, f_{\mathcal{M}'}(\mathcal{X}_{\mathbf{D_r}})  \stackrel{}{\longrightarrow} \mathcal{Y}_{\mathbf{D_r}}.
\end{equation}
Here, $\not \stackrel{}{\longrightarrow}$ means the mapping relationship no longer holds.

For IU, let there are a sequence of unlearning task, $\{\mathcal{T}_1,\mathcal{T}_2,\cdots,\mathcal{T}_{T}\}$ and $T$ is the number of unlearning tasks. And $\{\mathbf{D}_{f,t}, \mathbf{D}_{r,t}\}$ is the forgetting classes and remaining classes belonging to $\mathcal{T}_{t}$, where $\mathbf{D}_{r,t} = \mathbf{D} -  \cup_{i=1}^{t}\mathbf{D}_{f,i}$. Starting from $\mathcal{M}_{0}=\mathcal{M}$, the forgetting algorithm $\mathscr{F}$ iteratively processes till $t$-th unlearning task $\mathcal{T}_t$ to generate $\mathcal{M}_{t} = \mathscr{F}(\mathcal{M}_{t-1},\mathbf{D}_{f,t},\mathbf{D}_{r,t})$ which holds the followings:
\begin{equation}
f_{\mathcal{M}_{t}}(\mathcal{X}_{\mathbf{D}_{f,i}}) \not \stackrel{}{\longrightarrow} \mathcal{Y}_{\mathbf{D}_{f,i}}, f_{\mathcal{M}_{t}}(\mathcal{X}_{\mathbf{D}_{r,j}})  \stackrel{}{\longrightarrow} \mathcal{Y}_{\mathbf{D}_{r,j}},
\end{equation}

\vspace{-3mm}
\section{Methods}
\label{sec:methods}
\begin{figure*}
    % \vspace{-7mm}
    \centering
    \includegraphics[width=1\textwidth]{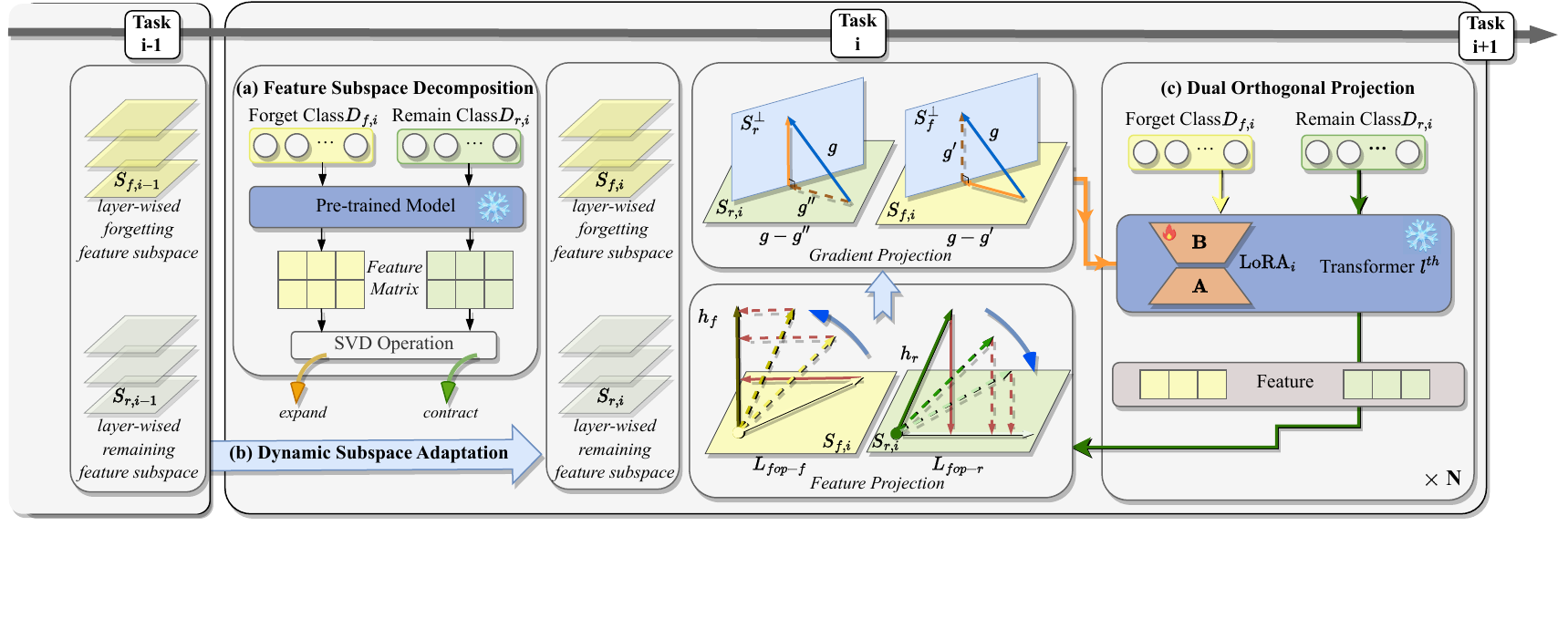}
    \vspace{-19mm}
    \caption{\textbf{Overview of FG-OrIU:} (a) For each unlearning task, we decompose the layer-wised- forgetting and remaining subspace separately (Sec. \ref{Feature_Subspace_Decomposition}). (b) LoRA is inserted at each layer, with constraints applied at both the feature and gradient levels for forgetting and remaining (Sec. \ref{Dual_Orthogonal_Projection}). (c) For new unlearning tasks, the forgetting subspace expands while the remaining contracts (Sec. \ref{Dynamic_Subspace_Adaptation}).}
    \vspace{-6mm}
    \label{fig:main_fig}
\end{figure*}

\vspace{-2mm}
In this section, we introduce \textbf{FG-OrIU}, a comprehensive framework for Incremental Unlearning (IU) in pre-trained models (PTM). By imposing orthogonal constraints on both feature representations and gradient updates, FG-OrIU ensures thorough removal of unwanted knowledge while preserving performance on remaining classes. An overview of the framework is shown in Figure \ref{fig:main_fig}.

% We propose FG-OrIU, a framework for Incremental Unlearning (IU) that operates through three core stages: (1) feature subspace decomposition to separate forgetting and retained class representations, (2) dual orthogonal projection to enforce feature suppression and gradient correction, and (3) dynamic subspace adaptation for sequential unlearning. As illustrated in Figure \ref{fig:main_fig }, FG-OrIU achieves irreversible forgetting by explicitly decoupling unwanted features while maintaining retention robustness, even under continuous deletion requests.

\vspace{-2mm}
\subsection{Feature Subspace Decomposition}
\label{Feature_Subspace_Decomposition}
\vspace{-2mm}
PTMs acquire implicit knowledge from pre-training data, benefiting downstream tasks during inference. Unlearning aims to selectively remove the influence of forgetting classes from the PTM \cite{cheng2024remaining}. Unlike explicit knowledge in traditional Knowledge Bases (KBs) \cite{roth2002probabilistic,surdeanu2014overview}, which can be directly edited, the knowledge in PTMs is implicit and intertwined across all parameters \cite{zhu2020modifying,cheng2024remaining,qin2024machine}. Effective unlearning requires precisely identifying feature spaces for forgetting and remaining. To achieve this, we decompose features into two subspaces: one capturing forgetting-class contributions and the other preserving remaining-class knowledge. Specifically, we perform singular value decomposition (SVD) on the feature matrix of forgetting classes and use the span of its dominant singular vectors to define the forgetting subspace. Similarly, we apply the same procedure to the remaining classes to construct its subspace. We then detail the process as follows.

Given a PTM, $\mathcal{M}$, we construct a representation matrix $\bm{R}_{f,i}=\left[\bm{x}_{1,i},\dots,\bm{x}_{n_{f_{i}},i}\right] \in \mathbb{R}^{n_{f_{i}} \times d}$ at each layer $l$, where columns correspond to representations of $n_{f_{i}}$ samples from the remaining classes $\mathbf{D_{f,i}}$, and $d$ is the output dimension\footnote{Here, we omit the notation $l$ for simplicity.}. SVD is applied to $\bm{R}_{f,i}=\bm{U}_{f,i}\bm{\Sigma}_{f,i}(\bm{V}_{f,i})^{T}$, followed by its $k$-rank approximation ${(\bm{R}_{f,i})}_{k}$ satisfying $||(\bm{R}_{f,i})_{k}||_{F}^{2} \geq \epsilon ||\bm{R}_{f,i}||_{F}^{2}$ for a threshold $\epsilon$. The subspace $\mathcal{S}_{f,i} = \text{span} \left\{\bm{B}_{f,i}\right\}$ is defining using the first $k$ vectors $\bm{B}_{f,i}=\left\{\bm{u}_{1,i}, \dots, \bm{u}_{k,i}\right\}$ from $\bm{U}_{f,i}$, capturing directions with the highest singular values. Similarly, the feature matrix $\bm{R}_{r,i}$ for remaining classes $\mathbf{D_{r,i}}$ defines the subspace $\mathcal{S}_{r,i} = \text{span} \left\{\bm{B}_{r,i}\right\}$. The two sets of layer-wised feature subspace, $\{\mathcal{S}_{f,i}^{1}, \dots,\mathcal{S}_{f,i}^{l}\}$ and $\{\mathcal{S}_{r,i}^{1}, \dots,\mathcal{S}_{r,i}^{l}\}$, represent the core part for forgetting and remaining classes, respectively, isolating features to be erased while preserving those to retain. By explicitly decomposing feature subspaces across all layers, we strengthen supervision signals for parameter updates and mitigate the risk of gradient obfuscation \cite{athalye2018obfuscated} in deep networks, where gradient inaccuracies caused by layer-wise nonlinear transformations could otherwise weaken the unlearning effect. This multi-layer constraint ensures consistent suppression of forgetting class features throughout the model hierarchy.

% We then detail the construction process, omitting the notation $l$ for simplicity.

\vspace{-2mm}
\subsection{Dual Orthogonal Projection}
\label{Dual_Orthogonal_Projection}
% \vspace{-2mm}
Prior to unlearning task $\mathcal{T}_i$, we freeze both the backbone and the classification head of the PTM and insert lightweight LoRA modules \cite{hu2022lora} at each layer. We then denote the model with LoRA inserted as $\mathcal{M}_{\theta}$, where $\theta$ represents the trainable parameters in LoRA. After unlearning, the updates in LoRA, $\theta^{*}$, will be merged into the initial weight of the backbone, $\mathcal{M}_{i} \overset{\theta^{*}}{\underset{\text{merge}}{\longrightarrow}} \mathcal{M}_{i+1}$.

Building on the feature subspaces $\mathcal{S}_{f}$ and $\mathcal{S}_{r}$ defined in Section \ref{Feature_Subspace_Decomposition}, we propose a dual orthogonal projection mechanism that operates through both forward feature alignment and backward gradient correction.

% Inspired by \cite{tu2025sdwv,yoon2024safree}, for forgetting classes, achieving thorough removal at the feature level requires that the features generated by the unlearned model, $\mathcal{M}_{\theta}$, deviate as much as possible from the feature subspace ${S}_{f}$. Meanwhile, we aim to ensure that the unlearned model's feature subspace for remaining classes aligns with ${S}_{r}$. The orthogonality constraint between ${S}_{f}$ and the unlearned features serves as a critical mechanism for facilitating class-specific forgetting: under a fixed classification head, features orthogonal to the original subspace ${S}_{f}$ induce maximal perturbations in class logits, as the inner product between orthogonal vectors diminishes to zero. This forces the model to suppress discriminative patterns associated with forgetting classes in its intermediate representations. By enforcing layer-wise orthogonality between feature directions and ${S}_{f}$, we systematically disrupt the parameter updates linked to class-specific information propagation across the network hierarchy, effectively erasing the model’s capacity to encode or reconstruct semantic attributes of the target classes.

\subsubsection{Feature Orthogonal Projection} 
Inspired by \cite{tu2025sdwv,yoon2024safree}, for forgetting classes, achieving thorough removal at the feature level requires that the features generated by the unlearned model, $\mathcal{M}_{\theta}$, deviate as much as possible from the feature subspace $\mathcal{S}_{f}$. Meanwhile, we aim to ensure that the unlearned model's feature subspace for remaining classes aligns with $\mathcal{S}_{r}$. The orthogonality constraint between $\mathcal{S}_{f}$ and the unlearned features serves as a critical mechanism for facilitating class-specific forgetting: under a fixed classification head, features orthogonal to the original subspace $\mathcal{S}_{f}$ induce maximal perturbations in class logits, as the inner product between orthogonal vectors diminishes to zero. This forces the unlearned model to suppress discriminative patterns associated with forgetting classes in its intermediate representations. To do so, we propose the following two regularization constraints.
% \begin{tcolorbox}[colback=gray!10, colframe=white, sharp corners, boxrule=0pt]\small
For forgetting classes, let $\bm{x_{f}}$ denote a batch of forgetting samples from $\mathbf{D_{f,i}}$, and $\bm{h}_f^{l}$ denote the $l$-th layer features gained with the unlearned model $\mathcal{M}_{\theta}$. For complete forgetting, we enforce:

\begin{align}
\label{eq1}
\vspace{-2mm}
\bm{h}_f^{l} \perp \mathcal{S}_{f,i}^l \quad \Rightarrow \quad \min ||\mathbf{P}_{f}^{l} \bm{h}_f^{l}||_F^2
\vspace{-2mm}
\end{align}
where $\mathbf{P}_{f}^{l}=\mathbf{B}_{f,i}^{l}{(\mathbf{B}_{f,i}^{l})}^{\top}$ is the projection matrix for the forgetting subspace $\mathcal{S}_{f,i}^{l}$ and $||\cdot||_{F}^{2}$ is the F-norm.
Eq.(\ref{eq1}) ensures that the features of $\mathbf{D_{f,i}}$ gained from the unlearned model $\mathcal{M}_{\theta}$ lie outside the original feature subspace gained from $\mathcal{M}$, erasing the unwanted information. Thus, we define the \textit{fop-f loss}:
\begin{align}
\label{loss:L_fop-f}
\vspace{-2mm}
\mathcal{L}_{fop-f} = \sum_{l=1}^{m} ||\mathbf{P}_{f}^{l}\bm{h}_f^{l}||_F^2.
\vspace{-4mm}
\end{align}
where $m$ denotes the total number of layers in the PTM.
For remaining classes, we follow previous works and apply the standard cross-entropy loss by constraining the model’s final output, as follows:
\begin{align} 
\label{loss:L_ce}
\mathcal{L}_{ce} = \mathcal{L}(\mathcal{X}_{\mathbf{D}_{r,1}}, \mathcal{Y}_{\mathbf{D}_{r,1}}), 
\end{align}
where $\mathcal{L}$ denotes the cross-entropy loss function. In addition, we also propose applying explicit constraint at the feature level for the remaining classes to better preserve the desired information. Specifically, $\bm{x_{r}}$ denote a batch of remaining samples from $\mathbf{D_{r,i}}$, and and $\bm{h}_r^{l}$ denote the $l$-th layer features gained with the unlearned model $\mathcal{M}_{\theta}$. For better preservation, we maintain feature fidelity through:
\begin{align}
\label{eq2}
\min ||\bm{h}_r^{l} - \mathbf{P}_{r}^{l} \bm{h}_r^{l}||_F^2
\end{align}
where $\mathbf{P}_{r}^{l}=\mathbf{B}_{r,i}^{l}{(\mathbf{B}_{r,i}^{l})}^{\top}$ is the projection matrix for the remaining feature subspace $\mathcal{S}_{r,i}^{l}$.

Eq.(\ref{eq2}) ensures that the features of $\mathbf{D_{r,i}}$ gained from the unlearned model $\mathcal{M}_{\theta}$, align with the orignal feature subspace gained from $\mathcal{M}$, maintaining the better preservation. Therefore, We draw the following \textit{fop-r} loss:
\begin{align}
\label{loss:L_fop-r}
\mathcal{L}_{fop-r} = \sum_{l=1}^{m} ||\bm{h}_r^{l} - \mathbf{P}_{r}^{l}\bm{h}_r^{l}||_F^2.
\end{align}

Finally, the total loss is:
\begin{align}
\label{eq:total_loss}
\mathcal{L}_{total} = \mathcal{L}_{ce} + \lambda_1 \mathcal{L}_{fop-f} + \lambda_2 \mathcal{L}_{fop-r},
\end{align}
where $\lambda_1$, $\lambda_2$ are the coefficients during training.

\subsubsection{Gradient Orthogonal Projection} 
To prevent parameter updates from reintroducing forgotten knowledge while preserving retained features, we design a gradient correction mechanism inspired by gradient projection techniques in multi-task learning \cite{yu2020gradient} and continual learning \cite{zhao2023rethinking,saha2021gradient}. Given the raw gradient $\bm{g_{l}}$ at layer $l$, we apply the following modifications:
\begin{align}
\label{eq:original approx unlearn steepest descent}
&\hat{\bm{g_{l}}}=\bm{g_{l}} - {\bm{g'_{l}}} - {\bm{g''_{l}}}\\
&=\bm{g_{l}} - \underbrace{\bm{g_{l}}(\bm{I} - \mathbf{B}_{f,i}^{l}{(\mathbf{B}_{f,i}^{l})}^{\top})}_{(term1: gop-f)} - \underbrace{\bm{g_{l}}\mathbf{B}_{r,i}^{l}{(\mathbf{B}_{r,i}^{l})}^{\top}}_{(term2:gop-r)}, \nonumber
\end{align}

\noindent
\textbf{Term (gop-f), $\bm{g'_{l}}$:} Removing the components of $\bm{g_{l}}$ that are orthogonal to $\mathcal{S}_{f,i}^{l}$ not only amplifies updates focusing on directions most critical for encoding forgetting class features, but also blocks parameter updates that could re-introduce removed knowledge.

\noindent
\textbf{Term (gop-r), $\bm{g''_{l}}$:} We aim to ensure that forgetting specific classes does not affect the remaining classes. Formally, for the unlearned model $f_{\mathcal{M}_{\theta}}$ and $\bm{x}_{r}$ from the remaining classes $\mathbf{D_{r,i}}$, the following equation holds: $f_{\mathcal{M}_{\theta}}(x_{r})=f_{\mathcal{M}}(x_{r})$. Specifically, we consider the input and output of layer $l$ as follows:
\begin{align}
\label{eq10}
\bm{W}_{l}\bm{x_{r}} + \Delta \theta_{l} \bm{x_{r}} &= \bm{W}_{l} \bm{x_r} \notag \\
\Delta \theta_{l} \bm{x_{r}} &= \bm{0}
\end{align}
where $\Delta \theta_{l}$ represents the change in the LoRA parameters inserted at layer $l$, which corresponds to its gradient. $\bm{W}_{l}$ is the parameter at layer $l$ from the pre-trained model, and these parameters remain frozen during unlearning.

Eq.(\ref{eq10}) indicates that if the update direction of the parameters, \ie, the gradient of the trainable LoRA parameters, lies in the complementary space of $\mathcal{S}_{r,i}^{l}$, \ie, $\bm{g_{l}} \perp \mathcal{S}_{r,i}^{l}$, then $\Delta \theta \bm{x_{r}} = \eta \bm{g_{l}} \bm{x_{r}} = \bm{0}$ (where $\eta$ is the learning rate), ensuring that the unlearned model’s performance on the remaining classes is not affected. Therefore, we further refine the gradient by removing its components along $\mathcal{S}_{r,i}^{l}$.

Finally, modified gradient $\hat{\bm{g}_{l}}$ serves the real gradient for parameter updates at the layer $l$.

\begin{table*}[t!]
\centering
\vspace{-5pt}
\resizebox{\textwidth}{!}{
\begin{tabular}{lrrrrrrrrrrrrrr}
\toprule
\multirow{2}{*}{Methods} &\multirow{2}{*}{\begin{tabular}[c]{@{}c@{}}Tunable\\ Ratio $\downarrow$ \end{tabular}} & \multicolumn{3}{c}{100-5} & \multicolumn{3}{c}{100-10} & \multicolumn{3}{c}{100-50} & \multicolumn{3}{c}{100-90} \\ 
\cmidrule(lr){3-5} \cmidrule(lr){6-8} \cmidrule(lr){9-11} \cmidrule(lr){12-14} 
 && $H \uparrow$ & $Acc_r \uparrow$ & $Acc_f \downarrow$ & $H \uparrow$ & $Acc_r \uparrow$ & $Acc_f \downarrow$ & $H \uparrow$ & $Acc_r \uparrow$ & $Acc_f \downarrow$ & $H \uparrow$ & $Acc_r \uparrow$ & $Acc_f \downarrow$ \\ \midrule
Pre-train & - & - & 73.85 &  70.88& -  & 73.81 & 72.74 & - & 72.45 & 74.88 & - & 72.32 & 73.86 \\
\midrule
% \multicolumn{5}{l}{\textcolor{orange}{\textit{(a) continual learning methods}}} \\
L2$^*$& 99.73\% & 67.13 & 66.96 & 3.58 & 67.74& 64.87&1.86& 64.39 & 56.61 & 0.24 & 52.45 & 40.68 & 0.04  \\
EWC$^*$  &99.73\%& 69.65 & 68.69 & {0.24} & 69.16 & 65.92 &{0.00}  & 62.48 & 53.61 & {0.00} & 44.41 & 31.75 & 0.00 \\
MAS$^*$ &99.73\% & 69.73 & 69.29 & 0.72 & 69.38 & 66.41 & 0.12 & 62.67 & 53.88 & {0.00} & 46.79 & 34.24 &{0.00}  \\
LwF &99.73\% & 67.95 &68.55&0.00 &70.08&67.62&0.00 &63.81&55.59&0.00 &61.25&52.32&0.00\\
DER &99.73\% &69.70 &68.55&0.00 &70.08&67.62&0.00&63.81&55.59&0.00&57.03&46.44&0.00\\
DER++ &99.73\% &69.70&68.56&0.00 &70.58&68.55&0.00&64.61&56.82&0.00&61.40&52.54&0.00\\
FDR  &99.73\%&70.31&69.74&0.00 &70.03&67.51&0.00&65.40&58.04&0.00&53.85&42.37&0.00\\
\midrule
% \multicolumn{5}{l}{\textcolor{orange}{\textit{(b) machine unlearning methods}}} \\
Retrain &100.00\%& 16.49&9.33&{0.00}&18.02&10.28&{0.00}&19.71&11.35&{0.00}&46.47&33.90&{0.00} \\
SCRUB &99.73\%& 67.78 & 64.94 & {0.00} & 68.26& 64.31&{0.00}& 65.82 & 58.71 & 0.00 & 16.11 & 9.04 & {0.00}  \\
SCRUB-S &99.73\%& 70.29 & {69.95} & {0.24} & 71.19 & {69.81} &{0.12}  & 57.49 & 51.20 & {9.34} & 17.53 & 9.94 & {0.00} \\
LIRF$^*$ &50.66\% & 25.56 & 67.67 & 55.13 & 26.35 & 65.83 & 56.26 & 47.13& 58.95 & {35.62} & 54.49 & 44.29 &{3.06}  \\
GS-LoRA  &\textbf{1.28\%}& {71.02} & {71.16} & 0.00 & {71.76} & {70.81} & {0.00}  & {71.29} & {68.05} & 0.02 & {73.71} & {73.56} & 0.00 \\
GS-LoRA++  &\textbf{1.28\%}& {71.12} & {71.36} & 0.00 & {72.04} & {71.35} & {0.00}  & {71.56} & {68.52} & 0.00 & {72.85} & {71.86} & 0.00 \\ 
\midrule
\rowcolor{gray!30} % 设置当前行颜色
\textbf{FG-OrIU} &\textbf{1.28\%}& \textbf{71.78} & {72.71} & {0.00} & \textbf{72.52} & {72.31} & {0.00}  & \textbf{74.18} & {73.51} & {0.00} & \textbf{75.49} & {77.15} & {0.00} \\ 
\bottomrule
\end{tabular}%
}
\vspace{-2mm}
\caption{{Static Machine Unlearning results for face recognition task on CASIA-Face100.} $Acc_r\ (\%)$ and $Acc_f\ (\%)$ are the accuracies of remaining and forgetting classes.
$^*$~denotes the original methods with a rehearsal buffer.
Retrain represents retraining the model using replay data and \textit{the training epoch is the same as other methods to ensure a fair comparison.} Pre-train denotes the results before forgetting.}
\label{tab:single-recog-caisa100_main}
\vspace{-6mm}
\end{table*}

\vspace{-2mm}
\subsection{Dynamic Subspace Adaptation}
\label{Dynamic_Subspace_Adaptation}
\vspace{-2mm}
When encountering a new unlearning task $\mathcal{T}_{i+1}$, our method only needs to update the two decomposed feature subspaces. Since the forgetting classes $\mathbf{D_{f,i+1}}$ in $\mathcal{T}_{i+1}$ are a subset of the remaining classes from the previous task $\mathcal{T}_{i}$, \ie, $\mathbf{D_{f,i+1}} \subseteq \mathbf{D_{r,i}}$, the feature subspaces can be efficiently adjusted. For simplicity, we omit the notion $l$ of layers.

We first construct a representation matrix $\bm{R}_{f,i+1}$ using only the forgetting classes $\mathbf{D_{f,i+1}}$. Before applying SVD and $k$-rank approximation, we remove common bases already present in $\mathcal{S}_{f,i}$ to ensure newly added bases are unique and orthogonal to existing ones:
% \begin{small}
    \begin{align}
    \label{subspace_update}
        {\hat{\bm{R}}}_{f,i+1} &= \bm{R}_{f,i+1} - \bm{B}_{f,i} \left(\bm{B}_{f,i}\right)^{T}  \left(\bm{R}_{f,i+1}\right) \\
        &= \bm{R}_{f,i+1} - \bm{R}_{f,i+1}^{proj}.
    \end{align}
% \end{small}
Next, we perform SVD on ${\hat{\bm{R}}}_{f,i+1}=\hat{\bm{U}}_{f,i+1}\hat{\bm{\Sigma}}_{f,i+1}(\hat{\bm{V}}_{f,i+1})^{T}$ and select $h$ new orthogonal from $\hat{\bm{U}}_{f,i+1}$. Then, the feature subspace of forgetting classes is expaneded as $\mathcal{S}_{f,i+1}=\text{span}\left\{\bm{B}_{f,i+1}\right\}$, where $\bm{B}_{f,i+1}=\left[\bm{B}_{f,i}; \bm{u}_{1}, \dots, \bm{u}_{h}\right]$ with $h$ new bases from $\hat{\bm{U}}_{f,i+1}$. And the feature subspace $\mathcal{S}_{r,i+1}$ of remaining classes is contracted to remove the overlapping features accordingly.

\vspace{-2mm}
\section{Experiment}
\label{sec:experiment}
\vspace{-2mm}
\subsection{Experimental Setup} 
\label{subsec:Experimental_Setup}
\noindent
\textbf{Datasets, Benchmarks and Pre-trained Models.}

\noindent Following \cite{zhao2024continual}, we evaluate FG-OrIU on Classification and Recognition tasks. For classification, we use ImageNet-100 \cite{russakovsky2015imagenet}, CUB-200 \cite{wah2011caltech}, and Omnibenchmark \cite{zhang2022benchmarking}, with a ViT-B/16 \cite{dosovitskiy2020image} pre-trained in PyTorch \cite{paszke2019pytorch}. For face recognition, we use CASIA-Face100 \cite{zhao2024continual}, a subset of CASIA-WebFace \cite{yi2014learning}, and MS-Celeb-100, sampled from \cite{guo2016ms}. The pre-trained model is the Face Transformer from \cite{zhong2021face}. We focus on class-unlearning scenarios with two benchmarks: (a) MU, varying numbers of classes are removed; (b) IU, 20 classes are removed per task over 4 tasks. \textbf{All setting is in the form of Y-X, which means \textit{experiments start from a PTM (Y classes originally) and forget X classes}.} More details about implementations are in Appendix \ref{sec:supp_impdetail}. 

\noindent
\textbf{Metrics.}
\noindent Following \cite{zhao2024continual}, we evaluate performance using average accuracy (\textit{Acc}) for both forgetting and remaining classes. The accuracy for forgetting classes should approach zero, while the accuracy for remaining classes should remain close to the original model's performance. Additionally, we adopt the H-Mean \cite{shibata2021learning} metric to assess overall performance after learning task $\mathcal{T}_t$, computed as:
\begin{small}
\begin{align} 
H\text{-}Mean^{(t)}=\frac{2Acc_r^{(t)} \cdot Drop^{(t)}}{Acc_r^{(t)}+Drop^{(t)}}. 
\end{align}
\end{small}
Here $Acc_r^{(t)}$ represents accuracy on the retained dataset after task $\mathcal{T}_t$, while $Drop^{(t)}=Acc_f^{(t-1)}-Acc_f^{(t)}$ measures the performance drop on forgetting classes before and after training. After unlearning task $\mathcal{T}_t$, we evaluate performance on all previously forgetting classes from tasks $\mathcal{T}_1,\mathcal{T}_2,\cdots,\mathcal{T}_{t-1}$.

\noindent
\textbf{Baselines.}
\noindent 
Following \cite{zhao2024continual}, we compare FG-OrIU against \textit{(a) continual learning (CL) methods}, including L2 regularization, EWC \cite{kirkpatrick2017overcoming}, MAS \cite{aljundi2018memory}, LwF \cite{li2017learning}, DER \cite{buzzega2020dark}, and FDR \cite{benjamin2018measuring}, as well as \textit{(b) machine unlearning (MU) methods} such as BAD-T \cite{chundawat2023can}, LIRF \cite{Ye2022LearningWR}, SCRUB/SCRUB-S \cite{kurmanji2023towards}, GS-LoRA/GS-LoRA++ \cite{zhao2024continual,zhao2025practical}, and retraining. To ensure forgetting occurs in the backbone, we freeze the classification head for all methods. In the retraining setup, we train a randomly initialized model using replay data, ensuring the number of training epochs matches that of other methods.

\begin{table*}[t!]
\centering
\small
%\resizebox{\textwidth}{!}{%
% \setlength{\tabcolsep}{3.0pt}
\resizebox{\textwidth}{!}{
\begin{tabular}{llllllllllllllll}
\toprule
\multirow{2}{*}{Methods} & \multicolumn{3}{c}{\textit{Task 1}(100-20)} & \multicolumn{4}{c}{\textit{Task 2}(80-20)} & \multicolumn{4}{c}{\textit{Task 3}(60-20)} & \multicolumn{4}{c}{\textit{Task 4}(40-20)} \\ \cmidrule(lr){2-4} \cmidrule(lr){5-8} \cmidrule(lr){9-12} \cmidrule(lr){13-16}
 & $H \uparrow$ & $Acc_r \uparrow$ & $Acc_f \downarrow$ & $H \uparrow$ & $Acc_r \uparrow$ & $Acc_f \downarrow$ & $Acc_o \downarrow$ & $H \uparrow$ & $Acc_r \uparrow$ & $Acc_f \downarrow$ & $Acc_o \downarrow$ & $H \uparrow$ & $Acc_r \uparrow$ & $Acc_f \downarrow$ & $Acc_o \downarrow$ \\ \midrule
Pre-train & - & 74.31 & 74.65 & - & 74.50 & 73.80 & - & - & 74.80 & 73.91 & - & - & 74.47 & 75.11 & - \\
\midrule
% \multicolumn{5}{l}{\textcolor{orange}{\textit{(a) continual learning methods}}} \\
L2$^*$ & 66.91 & 62.13 & 2.16 & 66.66 & 61.74 & 1.37 & 9.36 & 66.42 & 61.37 & 1.54 & 11.83 & 66.95 & 61.02 & 0.97 &8.00 \\
EWC$^*$ &67.71  & 61.95 & {0.00} & 67.71 & 62.55 & {0.00} & \textbf{{0.00}} & 67.09 &61.43  &0.00  & {0.10} & 68.02 & 62.14 &0.00  &{0.23}  \\
MAS$^*$ & 67.52 & 61.63 & 0.00 & 68.12 & 63.25 & 0.00 & \textbf{{0.00}}& 68.15 & 63.23 &  {0.00}& \underline{0.03}   & 67.70 & 61.61 &{0.00} &\textbf{0.00} \\
LwF &69.43&65.04&0.21&69.94&66.47&0.00&0.21&70.34&67.11&0.00&\textbf{{0.00}}&70.89&67.12&0.00&\underline{0.04} \\
DER &70.75&67.24&0.00&68.88&64.58&0.00&\textbf{{0.00}}&68.95&64.62&0.00&\textbf{{0.00}}&69.41&64.51&0.00&\textbf{0.00}\\
DER++ &70.01&65.90&0.00&68.99&64.77&0.00&\textbf{{0.00}}&69.96&66.42&0.00&\textbf{{0.00}}&69.85&65.28&0.00&\textbf{0.00}\\
FDR &68.29&62.92&0.00&67.39&62.01&0.00&\textbf{{0.00}}&69.16&64.99&0.00&\textbf{{0.00}}&72.51&70.08&0.00&\textbf{0.00}\\
\midrule
% \multicolumn{5}{l}{\textcolor{orange}{\textit{(b) machine unlearning methods}}} \\
Retrain &17.29&9.77&{0.00}&31.12&19.72&{0.00}&\textbf{{0.00}}&41.17&28.80&{1.72}&\textbf{{0.00}}&52.44&41.77&4.66&{0.09} \\
BAD-T & 68.90 & 63.97 & {0.00} & 69.56 & 65.95 & {0.21} & \textbf{{0.00}} & 70.21 & 66.87 & {0.00} & {0.34} & 72.71 & 70.46 & 0.00 &{0.04} \\
LIRF$^*$ & 30.59 & 64.46 & 54.60 & 34.05 & 62.03 & 50.34 & {43.50}& 44.56 & 62.53 &  {39.29}& {36.58}   & 40.36 & 62.62 &45.34 &27.96 \\
SCRUB & 70.39 & 66.59 & {0.00} & 70.55 & 67.85 & {0.32} & \textbf{{0.00}} & 71.01 & 68.33 & {0.00} & {0.44} & 73.36 & 71.68 & 0.00 &{0.05} \\
SCRUB-S &72.41  & 70.34 & {0.05} & {70.06}& {71.53} & {5.16} & {15.22} & {73.38}&73.09 &{0.24} & {15.81} & \underline{75.33} & 76.24 &{0.68}  &{6.47}  \\
% \rowcolor{yellow!50} % 另一行颜色
GS-LoRA & \underline{74.40} & {74.16} & 0.00 & \underline{73.59} & {73.37} &  0.00& \underline{0.05} & \underline{{74.36}} & {74.88} & 0.06 & \textbf{{0.00}} & {73.76} & {72.45} & 0.00 &{1.93} \\ 
% \rowcolor{yellow!50} %
GS-LoRA++ & {73.97} & {74.16} & 0.00 & {73.48} & {73.16} &  0.00& {0.10} & {74.20} & {74.51} & 0.00 & \textbf{{0.00} }& {75.23} & {75.36} & 0.00 &\textbf{0.00} \\
\midrule
\rowcolor{gray!30} % 设置当前行颜色
\textbf{FG-OrIU} & \textbf{74.45} & {74.25} & 0.00 & \textbf{74.64} & {75.50} &  0.00 & \textbf{0.00} & \textbf{75.49} & {77.14} & 0.00 & \textbf{{0.00} }& \textbf{77.71} & {80.51} & 0.00 &\textbf{0.00} \\
\bottomrule
\end{tabular}%
}
%}
\vspace{-3mm}
\caption{{Incremental Unlearning results for face recognition task on CASIA-Face100.} $Acc_o \ (\%)$ is the accuracy of old tasks, \ie, the accuracy on all previously forgetting classes in task $\{\mathcal{T}_1,\dots,\mathcal{T}_{t-1}\}$. With four tasks, each forgetting 20 classes, desired forgetting should consider both $Acc_r$ and $Acc_f$. We \textbf{bold} the best and \underline{underline} the second-best $H$ results.}
\label{tab:cl-recog-caisa100_main}
\vspace{-4mm}
\end{table*}

\begin{table*}[t!]
\centering
\small
%\resizebox{\textwidth}{!}{%
% \setlength{\tabcolsep}{3.0pt}
\resizebox{\textwidth}{!}{
\begin{tabular}{llllllllllllllll}
\toprule
\multirow{2}{*}{Methods} & \multicolumn{3}{c}{\textit{Task 1}(100-20)} & \multicolumn{4}{c}{\textit{Task 2}(80-20)} & \multicolumn{4}{c}{\textit{Task 3}(60-20)} & \multicolumn{4}{c}{\textit{Task 4}(40-20)} \\ \cmidrule(lr){2-4} \cmidrule(lr){5-8} \cmidrule(lr){9-12} \cmidrule(lr){13-16}
 & $H \uparrow$ & $Acc_r \uparrow$ & $Acc_f \downarrow$ & $H \uparrow$ & $Acc_r \uparrow$ & $Acc_f \downarrow$ & $Acc_o \downarrow$ & $H \uparrow$ & $Acc_r \uparrow$ & $Acc_f \downarrow$ & $Acc_o \downarrow$ & $H \uparrow$ & $Acc_r \uparrow$ & $Acc_f \downarrow$ & $Acc_o \downarrow$ \\ 
\midrule
Pre-train &- & 89.93 & 87.40 & - & 90.03 & 89.60 & - & - & 89.65 & 90.80 & - & - & 92.80 & 86.50 & - \\
\midrule
% \multicolumn{5}{l}{\textcolor{orange}{\textit{(a) continual learning methods}}} \\
L2$^*$ & 79.39 & 88.33 & 15.30 & 83.59 & 88.37 & 10.30 & 48.10 & 84.01 & 88.20 & 10.60 & 37.15 & 82.72 & 89.90 & 9.90 & 20.97 \\
EWC$^*$ &82.69 & 78.55 & 0.10 & 85.71 & 82.57 & 0.50 & 0.20 & 87.33 & 84.90 & 0.90 & 2.90 & 86.93 & 89.50 & 2.00 & 8.27  \\
MAS$^*$ & 80.44 & 74.50 & 0.00 & 81.51 & 74.77 & 0.00 & 0.60 & 82.18 & 75.05 & 0.00 & 0.60 & 81.93 & 77.90 & 0.10 & 1.77 \\
LwF &85.82 & 85.15 & 0.90 & \underline{87.26} & 85.40 & 0.40 & 2.40 & 87.35 & 84.15 & 0.00 & 2.35 & 86.05 & 85.70 & 0.10 & 1.07 \\
DER &82.88 & 89.73 & 10.40 & 84.22 & 90.57 & 10.90 & 50.80 & 86.36 & 91.20 & 8.80 & 42.00 & {88.52} & 94.70 & 3.40 & 24.90\\
DER++ &83.45 & 89.73 & 9.40 & 84.13 & 90.63 & 11.10 & 50.10 & \underline{87.50} & 91.30 & 6.80 & 40.40 & \textbf{89.24} & 94.80 & 2.20 & 20.80\\
FDR &29.51 & 17.78 & 0.50 & 36.21 & 22.70 & 0.10 & \textbf{0.00} & 40.79 & 26.30 & 0.00 & \textbf{0.00} & 47.91 & 33.50 & 2.40 & 1.27\\
\midrule
% \multicolumn{5}{l}{\textcolor{orange}{\textit{(b) machine unlearning methods}}} \\
Retrain &48.20 & 33.28 & 0.00 & 56.28 & 41.30 & 1.30 & \textbf{0.00} & 61.14 & 47.55 & 5.20 & \underline{0.20} & 67.23 & 60.60 & 11.00 & 0.53 \\
BAD-T & 74.74 & 65.29 & 0.00 & 78.09 & 69.21 & 0.00 & {0.10} & 76.65 & 66.32 & 0.00 & \textbf{0.00} & 76.25 & 68.17 & 0.00 & \textbf{0.00} \\
LIRF$^*$ & 52.72 & 64.28 & 42.71 & 61.05 & 69.83 & 35.36 & 32.15 & 60.96 & 60.32 & 29.18 & 30.91 & 58.51 & 68.46 & 35.41 & 21.04 \\
SCRUB & 75.87 & 67.03 & 0.00 & 76.23 & 66.33 & 0.00 & \textbf{0.00} & 75.46 & 64.55 & 0.00 & \textbf{0.00} & 77.56 & 70.30 & 0.00 & \textbf{0.00} \\
SCRUB-S &83.44 & 79.83 & 0.00 & 82.59 & 76.60 & 0.00 & \underline{0.10} & 79.81 & 71.20 & 0.00 & \textbf{0.00} & 78.23 & 71.40 & 0.00 & \underline{0.07} \\
% \rowcolor{yellow!50} %
GS-LoRA & {86.13} & {85.00} & {0.10} & \underline{87.33} & {85.27} & {0.10} & {1.10} & \underline{87.84} & {85.15} & {0.10} & {0.35} & 86.30 & {86.30} & {0.20} & \underline{0.07} \\ 
% \rowcolor{yellow!50} %
GS-LoRA++  & \underline{86.63} & {85.98} & {0.10} & 87.22 & {85.23} & {0.30} & {0.30} & 87.48 & {84.40} & {0.00} & {0.60} & 87.00 & {87.50} & {0.00} & \underline{0.07}\\ 
\midrule
\rowcolor{gray!30} % 设置当前行颜色
\textbf{FG-OrIU} & \textbf{87.01} & {86.73} & \textbf{0.10} & \textbf{88.45} & {87.43} & {0.10} & \underline{0.10} & \textbf{88.90} & {87.35} & {0.30} & \textbf{0.00} & \underline{89.15} & {92.10} & {0.10} & \textbf{0.00}\\ 

\bottomrule
\end{tabular}%
}
%}
\vspace{-3mm}
\caption{{Incremental Unlearning results for classification task on ImageNet100.}}
\label{tab:cl-class-imagenet100_main}
\vspace{-5mm}
\end{table*}

\begin{table}[t!]
  \centering
    \vspace{-2mm}
    \small
  \resizebox{\columnwidth}{!}{
  \begin{tabular}{lll|lll}
  \toprule
    % \multirow{2}{*}{Ablated Variants} & \multicolumn{2}{c}{\textcolor{orange}{\textit{Task 1}}} & \multicolumn{3}{c}{\textcolor{orange}{\textit{Task 2}}} \\
    \multirow{2}{*}{Variants} & \multicolumn{2}{c}{{\textit{Task 1(100-20)}}} & \multicolumn{3}{c}{{\textit{Task 2(80-20)}}} \\
    \cmidrule(lr){2-3} \cmidrule(lr){4-6} 
    & $Acc_r$ $\uparrow$ & $Acc_f$ $\downarrow$ & $Acc_r$ $\uparrow$ & $Acc_f$ $\downarrow$ & $Acc_o$ $\downarrow$ \\
    \midrule 
    Pre-train  & 74.30 & 74.60 & 74.50 & 73.80 & - \\
    \rowcolor{gray!30} % 设置当前行颜色
    \textbf{FG-OrIU}  & \textbf{74.25} & \textbf{0.00} & 75.50 & \textbf{0.00} & \textbf{0.00} \\
    \quad w/o \textit{fop-f} & 73.01 & 0.20 & 76.24 & 0.42 & 1.28 \\
    \quad w/o \textit{gop-f} & 73.92 & 0.10 & 76.58 & 4.37 & 0.82 \\  
    \quad w/o \textit{fop-r} & 73.92 & 0.05 & 74.98 & 0.21 & 0.15 \\
    \quad w/o \textit{gop-r} & 71.45 & 0.00 & 73.96 & 0.00 & 0.00 \\
     % && \gr{+1.72} & - & \bluegr{-0.49} & \bluegr{-8} \\
  \bottomrule
\end{tabular}
}
\vspace{-2mm}
\caption{Ablation studies on CASIA-Face100.}
\label{tab:ablation_table}
\vspace{-7mm}
\end{table}
% }

% \begin{figure}[!h]
%     \centering
%     \vspace{-2mm}
%     \includegraphics[width=1\linewidth]{./fi}
%         \vspace{-10mm}
%     \caption{Baseline: DualPrompt. Setting: IMR(Inc20Task10). The two figures respectively represent the evaluation results on the i-th task immediately after the completion of training for the i-th task.}
%     % \label{fig:performance_comparison}
%     \vspace{-6mm}
%     \label{label_pic_1}
%     % \vspace{-6mm}
% \end{figure}

% \setlength{\tabcolsep}{1.0mm}{
\begin{table}[t!]
  \centering
    \vspace{1mm}
    \small
  \resizebox{\columnwidth}{!}{
  \begin{tabular}{ll|llllll}
  \toprule
    % \multirow{2}{*}{Ablated Variants} & \multicolumn{2}{c}{\textcolor{orange}{\textit{Task 1}}} & \multicolumn{3}{c}{\textcolor{orange}{\textit{Task 2}}} \\
    % \cmidrule(lr){2-3} \cmidrule(lr){4-6} 
    Exps & Metirc & Pre-train & DER & FDR & SCRUB & GS-LoRA & \textbf{Ours} \\
    \midrule 
    \multirow{2}{*}{\textit{Exp1}} & SSIM & 0.94 & \cellcolor{yellow}0.60 & 0.34 & \cellcolor{yellow}0.75 & 0.39 & \cellcolor{gray!30}\textbf{0.01} \\
     & PSNR & 31.32 & \cellcolor{yellow}21.22 & 16.89 & \cellcolor{yellow}23.40 & 17.60 & \cellcolor{gray!30}\textbf{8.37} \\
     \midrule
    \multirow{2}{*}{\textit{Exp2}} & SSIM & 0.95 & 0.49 & \cellcolor{yellow}0.66 & 0.63 & 0.38 & \cellcolor{gray!30}\textbf{0.02} \\
     & PSNR & 31.15 & 19.17 & \cellcolor{yellow}22.15 & 21.12 & 17.32 & \cellcolor{gray!30}\textbf{8.72} \\
     \midrule
    \multirow{2}{*}{\textit{Exp3}} & SSIM & 0.93 & 0.53 & 0.49 & \cellcolor{yellow}0.60 & 0.41 & \cellcolor{gray!30}\textbf{0.01} \\
     & PSNR & 31.10 & 20.03 & 19.83 & \cellcolor{yellow}22.01 & 15.92 & \cellcolor{gray!30}\textbf{8.55} \\
  \bottomrule
\end{tabular}
}
\vspace{-2mm}
\caption{SSIM $\uparrow$ and PSNR(db) $\uparrow$ scores for the reconstructed images using DIP. Here, Pre-train serves as the upper bound.}
\label{tab:ssim_psnr}
\vspace{-3mm}
\end{table}
% }

\begin{figure}[t]
  \centering
   \includegraphics[width=1.0\linewidth]{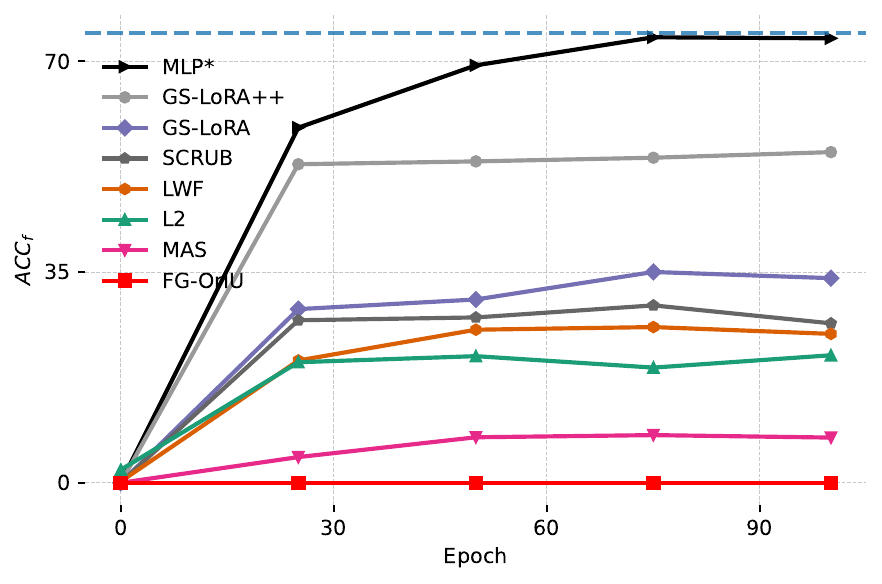}
   \vspace{-9mm}
   \caption{{Accuracy on forgetting classes when recovering on CASIA-Face100.} The \textcolor{blue}{line} (Pre-train) is the result before unlearning. The \textcolor{red}{line} (FG-OrIU) is Our method.}
   \label{fig:recover_face_figure}
   \vspace{-6mm}  
\end{figure}

\begin{figure*}[t]
  \centering
   \includegraphics[width=1.0\linewidth]{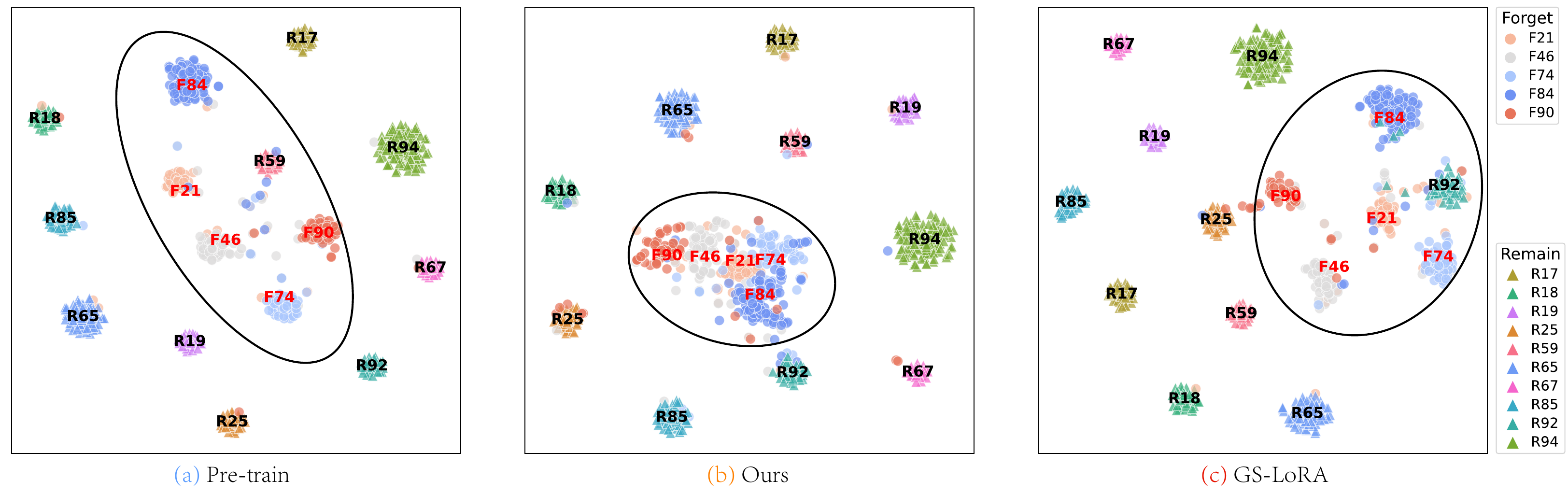}
   \vspace{-8mm}
    \caption{TSNE visualization with 5 forgetting classes and 10 remaining classes, using different model as feature extractor: \textcolor{blue}{(a)} the pre-trained model, \textcolor{orange}{(b)} the unlearned model through FG-OrIU, and \textcolor{red}{(c)} through GS-LoRA.}
  \label{fig:tsne}
   \vspace{-6mm}  
\end{figure*}

\vspace{-2mm} 
\subsection{Results and Comparison}
\vspace{-2mm} 
Table \ref{tab:single-recog-caisa100_main} presents the results for static Machine Unlearning task. FG-OrIU exhibits significantly degraded performance on forgetting classes while maintaining the performance of remaining classes. For example, in the settings where 50 and 90 classes are removed per task, FG-OrIU achieves improvements of 2.62\% and 2.64\% in $H$, and 4.99\% and 5.29\% in $ACC_{r}$ compared to GS-LoRA. Notably, FG-OrIU keeps the pre-trained model frozen and only trains the LoRA modules, enabling effective and efficient forgetting. Table \ref{tab:cl-recog-caisa100_main} and \ref{tab:cl-class-imagenet100_main} show the results for Incremental Unlearning task. FG-OrIU consistently outperforms all methods. Specifically, CL methods struggle to achieve effective forgetting on ImageNet100, as seen in DER, which retains high $ACC_f$ of 10.40\%, 10.90\%, 8.80\%, and 3.40\% across four tasks, respectively. On the other hand, MU methods fail to maintain their forgetting effect over sequential unlearning tasks. For example, LIRF and SCRUB-S show an increase in $ACC_o$ on CASIA-Face100, which becomes more pronounced as the sequence of unlearning tasks grows. GS-LoRA is specifically designed for IU. Compared to it, our method achieves better forgetting while also demonstrating superior performance in preserving the remaining classes, \ie, substantial improvements in $ACC_r$, particularly in long-sequence unlearning scenarios, achieving gains of 5.15\% and 4.6\% in the fourth unlearning task on two different datasets, respectively.

\vspace{-2mm} 
\subsection{Ablation Study}
\vspace{-2mm} 
Our FG-OrIU consists of four components, \ie, \textit{feature orthogonal projection for forgetting, feature orthogonal projection for remaining, gradient orthogonal projection for forgetting, and gradient orthogonal projection for remaining}, denoted as \textit{fop-f, fop-r, gop-f, and gop-r}. We independently remove each of these components to demonstrate their individual contributions. We observe that when the regularization term or gradient modification related to \textit{remaining} is removed, the performance on the remaining classes significantly declines. Meanwhile, when the regularization term or gradient modification related to \textit{forgetting} is removed, the performance on the forgetting classes shows a slight improvement. Thus, each component provides clear benefits in achieving a better trade-off between forgetting and remaining while preserving the old forgetting effect.
% Our FG-OrIU 包含4个部分，\ie，forward-disturbing, forward-denoisng, backward-disturbing, and backward-denoisng. We 在FG-OrIU中独立的将这四个组件删除，to展示每一个组件独立的贡献. 我们能够发现，当删除了关于denoising的正则项or gradient modification时，在remaining classes上的性能会出现明显的下降；同时当删除了关于disturbing的正则项or gradient modification时，关于forgetting classes的性能会出现轻微的上升。因此， each component yields clear benefits for 更好的达成forgetting and remaining 之间的trade-off 同时保持old forgetting effec.

% Overall, optimizing each component yields clear benefits, with all contributing to the robust gains of FG-OrIU. Besides, it is noteworthy that CPK do not reduce SSP, hence the performance improvement solely stemmed from the enhanced representational capacity of prompts. DGA not only integrates knowledge from multiple tasks into a single set of prompts, thereby enhancing the representational capacity, but importantly, the notable improvement in PRA is attributed to the reduction in the total amount of available sets during \emph{prompt retrieval}, thereby aiding PCL performance.

% \section{Discussion}
\vspace{-3mm}
\subsection{Discussion}
\vspace{-2mm}
\textcolor{black}{IU is designed to ensure strict privacy protection}. This raises several key questions: (a) Does the unlearned model still retain any private information? (b) Can the forgetting effect be reversed? and (c) How does unlearning impact the feature subspace? To address these questions, we design the following experiment.
\vspace{-2mm}
\subsubsection{Semantic Residual Analysis}
\vspace{-2mm}
We first investigate the semantic information contained in features after unlearning. We utilize Deep Image Prior (DIP) \cite{ulyanov2018deep} to reconstruct an image from the features obtained from the last layer of the unlearned model. Figure \ref{fig:figure1} presents the visualization of the reconstructed image, which allows us to infer the semantic information encoded in features. More results are provided in the Appendix \ref{app:dip}.

Additionally, we measure the SSIM and PSNR scores for the images generated by DIP from the unlearned model using different IU methods as a quantitative results. As shown in Table \ref{tab:ssim_psnr}, the SSIM values of the reconstructed images from the compared methods range between 0.35 and 0.75, with the highest being 0.75 for SCRUB on \textit{exp1}, while the PSNR values range between 16 and 24, with the highest being 23.40 for SCRUB on \textit{exp1}. This indicates that the reconstructed image achieves a high similarity with the original image, suggesting that the information in features has been partially preserved. In contrast, our method yields an SSIM of only 0.01 and a PSNR of 8.55, indicating a significant difference between the reconstructed and original images. This suggests that the semantic information at the feature level has been substantially obfuscated.
\vspace{-5mm}
\subsubsection{Forgetting Recoverable Analysis}
\vspace{-2mm}
Given an unlearned model obtained through different IU methods, we freeze its backbone and retrain a new classification head using the entire training dataset, which includes both forgetting and remaining classes. As illustrated in Figure \ref{fig:recover_face_figure}, MLP* represents a trivial unlearning method where the classification head's parameters for the forgetting classes are simply set to zero. Consequently, the underlying features remain intact, enabling effortless recovery of the forgotten knowledge. Additionally, existing IU methods primarily impose constraints on the final output rather than directly modifying feature representations. As a result, while features may be weakened to some extent, their core structure remains largely unchanged. Thus, these methods allow the forgotten knowledge to be recovered to 15\%–70\% of its original performance, which leads to \textit{superficial forgetting}. In contrast, our method achieves \textit{deep forgetting}, making recovery infeasible.
\vspace{-2mm}
\subsubsection{Feature Subspace Analysis}
% \vspace{-1mm}
To analyze the impact of direct constraints on features and gradients for forgetting and remaining classes, we conduct an experiment. We remove 20 classes while retaining 80, then extract features from 5 forgetting and 10 remaining classes and visualize them using t-SNE.

A key observation is that GS-LoRA causes some forgetting class samples to move closer to or even overlap with certain remaining class samples (e.g., \textcolor{red}{F21}, \textcolor{red}{F74} $\rightarrow$ \textcolor{black}{R92} in \textcolor{red}{c}). This suggests that merely applying constraints at the final output, resulting in degraded features from the forgetting classes entangling with those of the remaining classes. Consequently, this degradation negatively impacts the performance of the remaining classes. In contrast, our method enforces forgetting at both feature and gradient level, ensuring that features from the forgetting classes do not shift toward any remaining class but instead become linearly inseparable within the forgetting set. Notably, an unexpected yet beneficial side effect emerges: in the original PTM, certain remaining class samples (e.g., R59) were slightly entangled with forgetting classes (e.g., \textcolor{red}{F90} and \textcolor{red}{F46}). However, after applying our method, these classes become completely disentangled. This observation aligns with the results in Table \ref{tab:cl-recog-caisa100_main}, where the accuracy of the remaining classes improves after unlearning (e.g., 75.36 $\rightarrow$ 80.51), demonstrating that our method not only enhances forgetting effectiveness but also benefits the generalization of the remaining classes.

% \vspace{-2mm}
\section{Conclusion}
\label{sec:conclusion}
\vspace{-2mm}
In this work, we propose FG-OrIU for incremental unlearning in vision PTMs. By enforcing orthogonal constraints on both features and gradients, FG-OrIU achieves irreversible forgetting while maintaining performance on the remaining classes. Its dynamic subspace adaptation ensures robust handling of sequential deletion requests. Extensive experiments confirm its superiority. While FG-OrIU is designed for class-level unlearning, extending it to sample-level settings remains a promising direction for future work.
\section*{Acknowledgments} 
This work was supported in part by National Natural Science Foundation of China under Grant 62402430, 62206248, 62476238, Zhejiang Provincial Natural Science Foundation of China under Grant LQN25F020008, and Aeronautical Science Foundation of China 20240048076001. Qian Feng would like to extend his gratitude to Jiahang Tu and Hanbin Zhao from Zhejiang University for their discussions regarding the methods and experiments. Additionally, he is thankful to Mintong Kang from UIUC for his dedicated suggestions on improving the overall writing of the paper.
{

% \clearpage
% \newpage
    \small
    \bibliographystyle{ieeenat_fullname}
    \bibliography{main}
}

\clearpage
\setcounter{page}{1}
\maketitlesupplementary
\appendix

\noindent\textbf{Overview.} In this supplementary material, we provide the source code in the `FG-OrIU' folder and present additional experiments related to our method as follows:

In Section \ref{sec:supp_impdetail}, we provide detailed information about the datasets used, the benchmarks constructed separately for machine unlearning and incremental unlearning, the pre-trained models adopted along with their sources, and the hyperparameters used in our experiments.

In Section \ref{append:algho}, we provide the pseudocode of our FG-OrIU method.

In Section \ref{app:Extra_Experimental_Evaluations}, we present additional experimental results in the IU scenario, including results from the face recognition and image classification tasks. Furthermore, we provide additional visualization results using T-SNE.

In Section \ref{app:dip}, we provide a detailed introduction to Deep Image Prior, the technique used for feature reconstruction in this work, including its principles, implementation, and additional visualization results.

% In this section, we discuss the detailed implementation in FG-OrIU, including the introduction of datasets, selection or pre-training of the pre-trained models, compared methods, and hyper-parameters.

\section{Implementation Details} 
\label{sec:supp_impdetail}
In this section, we discuss the detailed implementation in FG-OrIU, including the introduction of datasets, selection or pre-training of the pre-trained models, and hyper-parameters.

\subsection{Datasets}

In this section, we introduce the datasets used in this paper. The details of the five adopted datasets are listed in Table~\ref{tab:supp_dataset}. Specifically, CASIA-Face100 is a subset of CASIA-WebFace \cite{yi2014learning}, provided by \cite{zhao2024continual}. Following a similar approach, we sample a subset with 100 face identities from the large-scale face dataset MS-Celeb-1M \cite{guo2016ms}. Additionally, CUB200, ImageNet-100, and OmniBenchmark are widely used benchmark datasets for continual learning, frequently adopted in studies such as \cite{rebuffi2017icarl,dong2022federated_FCIL,10323204, wu2024soediff, wu2024mote, wu2024individual, pan2024towards, wu2024self, tu2025texttoucher, tu2025driveditfit, wang2025unleashing, wang2025efficiently, wang2024belm, wang2024gad, feng2024lw2g, feng2025pectp}.

\begin{table*}
	\caption{Introduction about benchmark datasets. CASIA-Face100 and MS-Celeb-100 are for Face Recognition Task. CUB-200, ImageNet-100, and OmniBenchmark are for Image Classification Task.}
	\label{tab:supp_dataset}
	\centering
	%	\resizebox{1.0\linewidth}{!}{
		\begin{tabular}{lcccccc}
			\toprule
			Dataset & \# training instances & \# testing instances & \# Classes& Link \\ \midrule
			CASIA-Face100 & 35713 & 8984 & 100 & \href{https://github.com/bjzhb666/GS-LoRA}{\nolinkurl{Link}}\\
			MS-Celeb-100 & 10,517 & 2684 & 100 & \href{https://github.com/deepinsight/insightface/tree/master/challenges/iccv19-lfr}{\nolinkurl{Link}}\\
                \midrule
			CUB-200 & 9,430 & 2,358 & 200 & \href{https://www.vision.caltech.edu/datasets/cub_200_2011/}{\nolinkurl{Link}}\\
			ImageNet-100 & 130,000 & 5000 & 100 & \href{https://www.kaggle.com/datasets/ambityga/imagenet100}{\nolinkurl{Link}}\\
			% \midrule
			OmniBenchmark & 89,668 & 5,983 & 300 & \href{https://github.com/ZhangYuanhan-AI/OmniBenchmark}{\nolinkurl{Link}}\\
			\bottomrule
		\end{tabular}
		%	}
\end{table*}

\subsection{Pre-Trained Models}

Before performing unlearning, we first need to obtain a pre-trained model. Therefore, to validate that our method can be flexibly scaled and effectively applied across models of different parameter sizes for incremental unlearning, we evaluate multiple pre-trained models, detailed as follows:

\noindent
\textbf{Face Recognition Task on CASIA-Face100 Dataset}

Following \cite{zhao2024continual}, we train a face transformer with six transformer blocks from scratch on the CASIA-Face100 dataset and use it as the original model.

\noindent
\textbf{Face Recognition Task on MS-Celeb-100 Dataset}

We directly utilize the checkpoints released by \cite{zhong2021face}, specifically a face transformer with 20 transformer blocks. This model was pre-trained on the large-scale face dataset MS-Celeb-1M \cite{guo2016ms}, and the released checkpoint includes a classification head with a dimensionality of 93,431. We then freeze the backbone and perform a linear probe, retraining a classification head with a dimensionality of 100 on the MS-Celeb-100 dataset to obtain the original model.

\noindent
\textbf{Image Classification Task on ImageNet-100 Dataset}

Following \cite{zhao2024continual}, we use the pre-trained ViT-B/16 model \cite{dosovitskiy2020image}, which was trained in PyTorch \cite{paszke2019pytorch} on ImageNet-21K \cite{deng2009imagenet}, as the original model.

\noindent
\textbf{Image Classification Task on CUB-200 dataset}

We utilize the pre-trained ViT-B/16 model, freeze its backbone, and perform a linear probe by retraining a classification head with a dimensionality of 200 on the CUB-200 dataset to obtain the original model.

\noindent
\textbf{Image Classification Task on OmniBenchmark Dataset}

Similarly, we use the pre-trained ViT-B/16 model, freeze its backbone, and perform a linear probe by retraining a classification head with a dimensionality of 300 on the OmniBenchmark dataset to obtain the original model.

% \subsection{Comparing Methods}
% xxxxxx

\subsection{Implementations and Hyper-Parameters}

We follow the implementation of GS-LoRA\footnote{\url{https://github.com/bjzhb666/GS-LoRA}} \cite{zhao2024continual} to re-implement all the compared methods, including L2*, EWC*, MAS*, LwF, DER, DER++, FDR, Retrain, BAD-T, LRIF*, SCRUB, SCRUB-S, GS-LoRA, and GS-LoRA++. To ensure a fair comparison, we strictly follow the implementation details provided in \cite{zhao2024continual}, including data augmentation strategies and random seed settings. Following \cite{zhao2024continual, Wang_2022_CVPR}, all results are averaged over five runs with different seeds (42, 288, 488, 688, 1337). We use 1 4090 GPU for experiments in Face recognition Task and 1 A800 GPU for experiments in Image Classification Task.

\begin{figure*}[!h]
    \centering
    % \vspace{-2mm}
    \begin{subfigure}{1\linewidth}
        \centering
        \includegraphics[width=\linewidth]{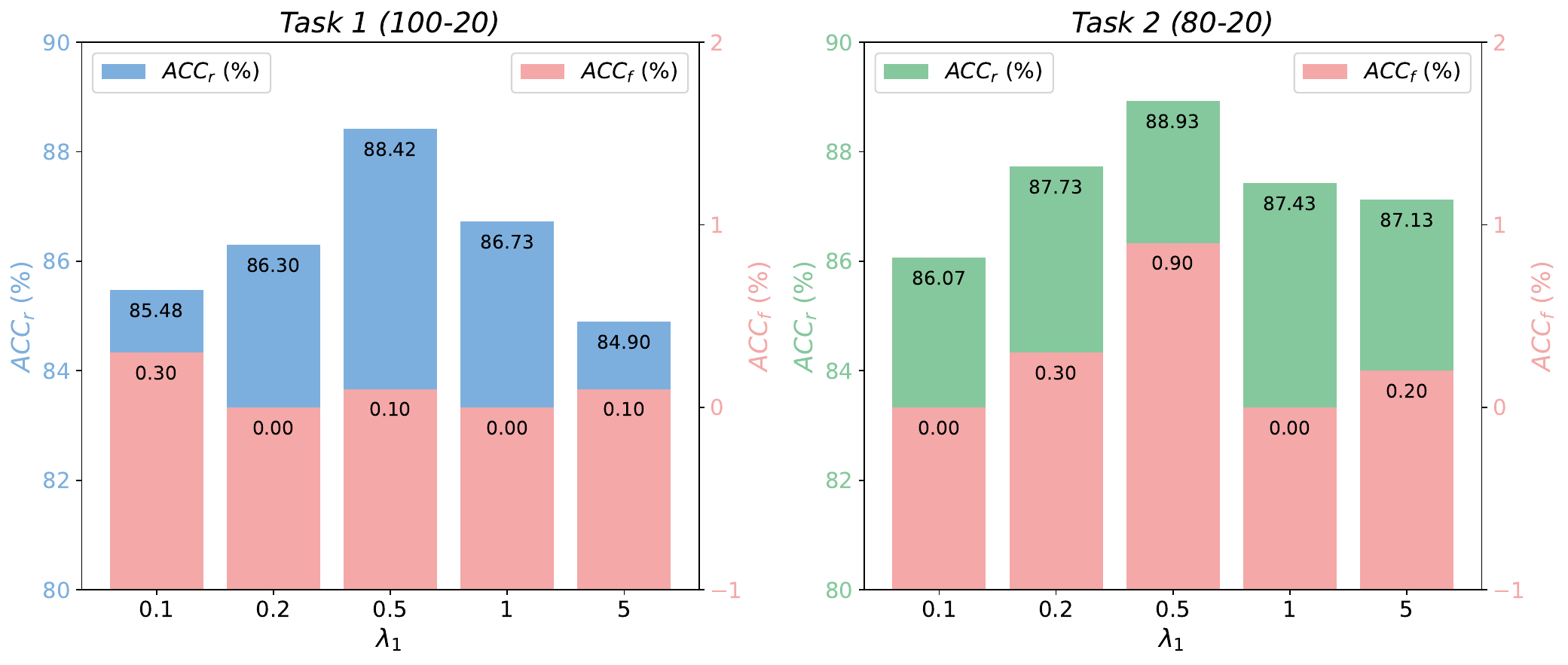}
        \caption{Ablations on $\lambda_1$ condition with $\lambda_2=1$.}
        \label{fig:ablation_1}
    \end{subfigure}
    \hfill
    \begin{subfigure}{1\linewidth}
        \centering
        \includegraphics[width=\linewidth]{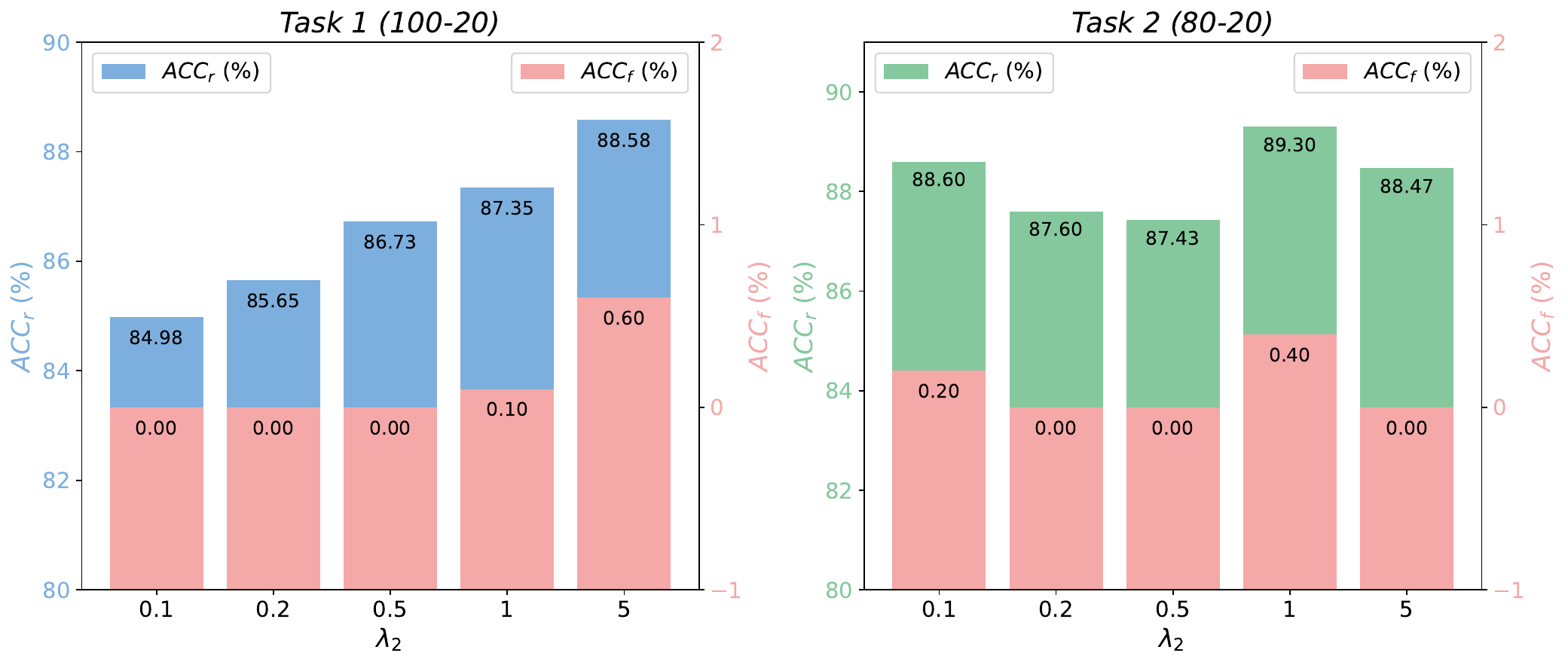}
        \caption{Ablations on $\lambda_2$ condition with $\lambda_1=0.2$.}
        \label{fig:ablation_2}
    \end{subfigure}
    % \vspace{-4mm}
    \caption{Ablation studies on $\lambda_1$ and $\lambda_2$ on ImageNet100 of Incremental Unlearning.}
    \label{fig:ablation_all}
    % \vspace{-6mm}
\end{figure*}

Our FG-OrIU method involves three hyperparameters: $\lambda_{1}$, $\lambda_{2}$, and $\epsilon$. The first two coefficients control the orthogonal constraints applied at the feature level. To analyze the impact of these coefficients in Eq.~(\ref{eq:total_loss}) on both forgetting and retention, we conduct experiments with different values of $\lambda_{1}$ and $\lambda_{2}$. The results are presented in Figure \ref{fig:ablation_all}.

We vary $\lambda_{1}$ across \{0.1, 0.2, 0.5, 1, 5\} and find that the model achieves optimal performance when $\lambda_{1} = 1$. Similarly, we experiment with $\lambda_{2}$ values of \{0.1, 0.2, 0.5, 1, 5\} and observe that our method attains the highest accuracy when $\lambda_{2} = 0.2$.

The parameter $\epsilon$ serves as a threshold for both constructing the feature subspace and its dynamic adaptations. Following    \cite{zhao2023rethinking,saha2021gradient}, we set $\epsilon = 0.99$ as the default value.

\section{Extra Experimental Evaluations} 
\label{app:Extra_Experimental_Evaluations}

\subsection{Additional Results on Computational cost and memory overhead}

FG-OrIU is designed for efficiency. Both the SVD step \ref{Feature_Subspace_Decomposition} and dynamic subspace updates \ref{subspace_update} are performed \textbf{only once per task}. Profiling with \textit{ptflops} shows the extra computational cost is only $\approx$1.5 GFLOPs per layer, compared to $\approx$17 GFLOPs per layer for ViT-B/16.

In terms of memory, we provide detailed subspace statistics using the 6-layer Face-Transformer. In Task 1, the number of bases per layer is [13,33,49,55,27,19] for the forgetting subspace and [14,35,55,71,65,69] for the remaining subspace; in Task 4, [14,35,55,71,64,67] (increases) and [14,33,49,57,34,20] (decreases). Each base is a 768-dim float32 vector, requiring only $\approx$0.2–0.3 MB per layer,far smaller than storing raw features/activations.

To further assess practical impact, we compare wall-clock Runtime and peak GPU memory with baseline methods. As shown in Table~\ref{tab2}, FG-OrIU remains lightweight with moderate Runtime and GPU usage, while achieving the highest performance with a notable gain over all baselines.

\begin{table}[!h]     
\centering
\small
%\resizebox{\textwidth}{!}{%
% \setlength{\tabcolsep}{3.0pt}
% \vspace{-13pt}
\resizebox{0.47\textwidth}{!}{
\begin{tabular}{ccccc}
\toprule
Metric & SCRUB & DER++  & GS-LoRA   & FG-OrIU \\
Runtime (s) / GPU Memory (MiB) & 1900/3753 & 2100/9611 & 2700/6299  & \cellcolor{gray!30}4000/6371 \\
Performance (\textit{H-Mean} via Eq.13) & 73.36 & 69.85 & 73.76  & \cellcolor{gray!30}77.71 \\
\bottomrule
\end{tabular}%
}
%}
% \vspace{-11pt}
\caption{Runtime and GPU memory comparison on A800 GPU.}
\label{tab2}
% \vspace{-15pt}
\end{table}

\subsection{Additional Comparison under Identical Tunable Ratio}

A lower tunable ratio enhances parameter efficiency and mitigates interference with remaining-class representations, thereby better preserving retention performance. To further investigate this effect, we conducted additional experiments under a fixed tunable ratio of 1.28\%, consistent with GS-LoRA and FG-OrIU, where the backbone is frozen and only the LoRA modules are trainable. As shown in Table~\ref{tab4}, although reducing the tunable ratio improves performance on $\mathbf{D_r}$, it significantly impairs forgetting effectiveness on $\mathbf{D_f}$. Existing methods struggle to balance this trade-off. In contrast, our approach achieves a more favorable balance between forgetting and retention, while maintaining high parameter efficiency.

\begin{table*}[!h]
    \centering
    \small
    %\resizebox{\textwidth}{!}{%
    % \setlength{\tabcolsep}{3.0pt}
    % \vspace{-13pt}
    \resizebox{\textwidth}{!}{
    \begin{tabular}{llllllllllll}
    \toprule
    \multirow{2}{*}{Method} & \multicolumn{2}{c}{\textit{Task 1}(100-20)} & \multicolumn{3}{c}{\textit{Task 2}(80-20)} & \multicolumn{3}{c}{\textit{Task 3}(60-20)} & \multicolumn{3}{c}{\textit{Task 4}(40-20)} \\ \cmidrule(lr){2-3} \cmidrule(lr){4-6} \cmidrule(lr){7-9} \cmidrule(lr){10-12}
     & $Acc_r \uparrow$ & $Acc_f \downarrow$ & $Acc_r \uparrow$ & $Acc_f \downarrow$ & $Acc_o \downarrow$  & $Acc_r \uparrow$ & $Acc_f \downarrow$ & $Acc_o \downarrow$ & $Acc_r \uparrow$ & $Acc_f \downarrow$ & $Acc_o \downarrow$ \\ \midrule
     \textbf{EWC*} & {71.60} & 11.67 & {73.37} & 13.88 & 9.97 & {73.52} & 7.75 & 5.64 & 74.52 & 4.48 & 1.42 \\
     \textbf{DER++} & {71.92} & 13.11 & 73.37 & 15.46 & 15.21 & 74.15 & 7.04 & 9.72 & {75.47} & 7.44 & 2.23 \\
     \textbf{GS-LoRA} & {74.16} & 0.00 & {73.37} &  0.00 & 0.05 & {74.88} & 0.06 & \textbf{{0.00}} & {72.45} & 0.00 & 1.93 \\
    \rowcolor{gray!30} % 设置当前行颜色
    \textbf{FG-OrIU} & {74.25} & 0.00 & {75.50} &  0.00 & \textbf{0.00} & {77.14} & 0.00 & \textbf{{0.00} } & {80.51} & 0.00 &\textbf{0.00} \\
    \bottomrule
    \end{tabular}%
    }
    %}
    % \vspace{-11pt}
    \caption{Extended results for Table \ref{tab:cl-recog-caisa100_main} \textbf{under 1.28\% tunable ratio}.}
    \label{tab4}
    % \vspace{-15pt}
\end{table*}

\subsection{Additional Results on Incremental Unlearning}
In this section, we conduct experiments to evaluate the scalability of FG-OrIU. Specifically, we perform incremental unlearning on the CUB-200 dataset, MS-Celeb-100, and OmniBenchmark using pre-trained models.
On CUB-200, we set up four unlearning tasks, each with 20 classes designated as forgetting classes.
On MS-Celeb-100, we similarly set up four unlearning tasks, each with 20 forgetting classes.
On OmniBenchmark, we set up four unlearning tasks, but each task contains 50 forgetting classes.
For a fair comparison, we re-implemented all baseline methods, and the corresponding results are presented in Tables 1, 2, and 3, respectively.

\begin{table*}[t!]
\centering
\small
%\resizebox{\textwidth}{!}{%
% \setlength{\tabcolsep}{3.0pt}
\resizebox{\textwidth}{!}{
\begin{tabular}{lllllllllllll}
\toprule
\multirow{2}{*}{Methods} & \multicolumn{2}{c}{\textit{Task 1}(100-20)} & \multicolumn{3}{c}{\textit{Task 2}(80-20)} & \multicolumn{3}{c}{\textit{Task 3}(60-20)} & \multicolumn{3}{c}{\textit{Task 4}(40-20)} \\ \cmidrule(lr){2-3} \cmidrule(lr){4-6} \cmidrule(lr){7-9} \cmidrule(lr){10-12}
 & $Acc_r \uparrow$ & $Acc_f \downarrow$ & $Acc_r \uparrow$ & $Acc_f \downarrow$ & $Acc_o \downarrow$ & $Acc_r \uparrow$ & $Acc_f \downarrow$ & $Acc_o \downarrow$ & $Acc_r \uparrow$ & $Acc_f \downarrow$ & $Acc_o \downarrow$ \\ \midrule
Pre-train & {99.53} & 99.61 & {99.63} & 99.25 & {-} & {99.45} & 100.0 & {{-}} & {99.30} & 99.63 & - \\
\midrule
% \multicolumn{5}{l}{\textcolor{orange}{\textit{(a) continual learning methods}}} \\
L2$^*$ & {98.43} & 0.19 & {98.58} & 0.74 & {23.71} &   {98.18} & 0.76 & {{25.52}} & {97.69} & 0.37 &{8.52} \\
EWC$^*$ & {99.44} & 0.00 & {99.32} & 0.00 & {0.00} &   {99.55} & 0.00 & {{0.00}} & {98.93} & 0.00 &{0.00} \\
MAS$^*$ & {99.44} & 0.00  & {99.38} & 0.00 & {0.00} &   {99.45} & 0.00 & {{0.00}} & {98.93} & 0.00 &{0.00} \\
LwF & {98.80} & 0.00  & {99.51} & 0.00 & {0.00} &   {99.64} & 0.00 & {{0.00}}   & {99.47} & 0.00 &{0.00} \\
DER & {99.72} & 0.00 & {99.69} & 0.00 & \underline{0.00}  & {99.90} & 0.00 & {{0.00}}  & {99.46} & 0.00 &{0.00} \\ 
DER++ & {99.67} & 0.00 & {99.69} & 0.00 & \underline{0.00}  & {99.63} & 0.00 & {{0.00}}  & {99.64} & 0.00 &{0.00} \\ 
FDR & {52.58} & 0.00 &  {28.84} & 0.00 & {0.00} &  {42.49} & 0.00 & {{0.00}}   & {53.28} & 0.00 &{0.00} \\
\midrule
% \multicolumn{5}{l}{\textcolor{orange}{\textit{(b) machine unlearning methods}}} \\
Retrain & {38.27} & 0.00 & {50.06} & 1.11 & {0.00} & {57.32} & 11.09 & {{0.00}}  & {64.12} & 25.75 &{1.07} \\
BAD-T & {99.13} & 0.00 & {99.62} & 0.00 & {0.00} & {99.58} & 0.00 & {{0.00}} & {99.77} & 0.00 &{0.00} \\
LIRF$^*$ & {86.35} & 81.64 & {81.81} & 74.03 & {55.64}  & {76.71} & 56.60 & {{40.87}}  & {73.00} & 63.99 &{25.05} \\
SCRUB & {99.58} & 0.00  & {99.14} & 0.00 & {0.00}  & {98.27} & 0.00 & {{0.00}}  & {97.87} & 0.00 &{0.00} \\
SCRUB-S & {99.49} & 0.00  & {99.57} & 0.00 & {0.00}  & {99.09} & 0.00 & {{0.00}}  & {98.93} & 0.00 &{0.00} \\
% \rowcolor{yellow!50} % 另一行颜色
GS-LoRA & {99.40} & 0.00  & {99.44} & 0.00 & {0.00}   & {99.09} & 0.00 & {{0.00}} & {99.12} & 0.00 &{0.00} \\ 
% \rowcolor{yellow!50} %
GS-LoRA++ & {99.35} & 0.00 & {99.45} & 0.00 & {0.00} & {99.00} & 0.00 & {{0.00}}  & {99.29} & 0.00 &{0.00} \\ 
\midrule
\rowcolor{gray!30} % 设置当前行颜色
FG-OrIU & {99.80} & 0.00   & {99.71} & 0.00 & {0.00}  & {99.88} & 0.00 & {{0.00}}  & {99.70} & 0.00 &{0.00} \\
\bottomrule
\end{tabular}%
}
%}
% \vspace{-2mm}
\caption{{Incremental Unlearning results for face recognition task on MS-Celeb-100.} With four tasks, each forgetting 20 classes.}
\label{tab:cl-recog-caisa100_main_ms100}
% \vspace{-3mm}
\end{table*}

\begin{table*}[t!]
    \centering
    \small
    %\resizebox{\textwidth}{!}{%
    % \setlength{\tabcolsep}{3.0pt}
    \resizebox{\textwidth}{!}{
    \begin{tabular}{lllllllllllll}
    \toprule
    \multirow{2}{*}{Methods} & \multicolumn{2}{c}{\textit{Task 1}(100-20)} & \multicolumn{3}{c}{\textit{Task 2}(80-20)} & \multicolumn{3}{c}{\textit{Task 3}(60-20)} & \multicolumn{3}{c}{\textit{Task 4}(40-20)} \\ \cmidrule(lr){2-3} \cmidrule(lr){4-6} \cmidrule(lr){7-9} \cmidrule(lr){10-12}
     & $Acc_r \uparrow$ & $Acc_f \downarrow$ & $Acc_r \uparrow$ & $Acc_f \downarrow$ & $Acc_o \downarrow$ & $Acc_r \uparrow$ & $Acc_f \downarrow$ & $Acc_o \downarrow$ & $Acc_r \uparrow$ & $Acc_f \downarrow$ & $Acc_o \downarrow$ \\ \midrule
    Pre-train  & {62.60} & 66.55 & {61.53} & 61.25 & {-} & {61.67} & 63.56 & {{-}} & {61.49} & 63.92 &{-} \\
    \midrule
    % \multicolumn{5}{l}{\textcolor{orange}{\textit{(a) continual learning methods}}} \\
    L2$^*$ &  {64.17} & 0.00 &  {64.31} & 3.80 & {15.70} &  {68.05} & 4.75 & {{26.29}} &  {64.97} & 2.58 &{21.42} \\
    EWC$^*$ &  {65.30} & 0.00 &  {65.12} & 0.00 & {3.75} &  {65.61} & 0.00 & {{3.44}} &  {65.32} & 0.00 &{3.06} \\
    MAS$^*$ &  {64.30} & 0.00 &  {63.82} & 0.00 & {0.00} &  {63.15} & 0.00 & {{0.25}} &  {62.47} & 0.00 &{0.64} \\
    LwF &  {66.51} & 0.00 &  {68.62} & 0.00 & {0.00} &  {68.61} & 0.18 & {{0.17}} &  {69.34} & 0.00 &{0.17} \\
    DER  & {42.59} & 3.75 &  {26.95} & 2.08 & {7.00} &  {14.99} & 1.76 & {{5.50}} &  {11.60} & 0.69 &{7.85} \\ 
    DER++  & {57.37} & 2.56 &  {60.19} & 1.90 & {4.26} &  {59.87} & 1.93 & {{3.44}} & {60.63} & 0.69 &{6.29} \\ 
    FDR &  {8.00} & 2.21 &  {11.32} & 4.15 & {2.05} &  {14.94} & 1.93 & {{2.57}} &  {18.79} & 1.71 &{2.31} \\
    \midrule
    % \multicolumn{5}{l}{\textcolor{orange}{\textit{(b) machine unlearning methods}}} \\
    Retrain &  {14.27} & 0.00 &  {13.60} & 0.00 & {0.00} &  {14.08} & 0.00 & {{0.00}}  & {14.65} & 0.00 &{0.00} \\
    BAD-T &  {44.53} & 0.00 &  {45.90} & 3.10 & {0.00} &  {43.97} & 2.90 & {{0.00}} &  {35.44} & 2.60 &{0.00} \\
    LIRF$^*$ &  {34.67} & 26.70 & 35.29 & 25.43 & {0.00} & {31.09} & 18.47 & {{0.00}} &  {30.28} & 11.08 &{0.00} \\
    SCRUB &  {48.58} & 0.00 &  {47.86} & 4.15 & {0.00} &  {46.45} & 26.41 & {{0.00}} & {46.44} & 29.90 &{0.75} \\
    SCRUB-S  & {67.87} & 0.00 &  {67.52} & 0.00 & {0.00} &  {66.05} & 25.70 & {{0.00}} &  {66.09} & 51.89 &{5.20} \\
    % \rowcolor{yellow!50} % 另一行颜色
    GS-LoRA  & {66.61} & 0.00 & {68.06} & 0.00 & {0.00}  & {69.89} & 0.17 & {{0.00}} & {70.78} & 0.00 &{0.05} \\ 
    % \rowcolor{yellow!50} %
    GS-LoRA++  & {66.99} & 0.00 & {68.79} & 0.00 & {0.00} & {68.76} & 0.00 & {{0.00}} & {70.23} & 0.00 &{0.05} \\ 
    \midrule
    \rowcolor{gray!30} % 设置当前行颜色
    FG-OrIU & {68.70} & 0.00 & {69.10} & 0.00 & {0.00} & {70.02} & 0.00 & {{0.00}} & {71.03} & 0.00 &{0.00} \\
    \bottomrule
    \end{tabular}%
    }
    %}
    \vspace{-2mm}
    \caption{{Incremental Unlearning results for image classification task on CUB-200.} With four tasks, each forgetting 20 classes.}
    \label{tab:cl-recog-caisa100_main_cub200}
    \vspace{-3mm}
\end{table*}

\begin{table*}[t!]
    \centering
    \small
    %\resizebox{\textwidth}{!}{%
    % \setlength{\tabcolsep}{3.0pt}
    \resizebox{\textwidth}{!}{
    \begin{tabular}{lllllllllllll}
    \toprule
    \multirow{2}{*}{Methods} & \multicolumn{2}{c}{\textit{Task 1}(100-20)} & \multicolumn{3}{c}{\textit{Task 2}(80-20)} & \multicolumn{3}{c}{\textit{Task 3}(60-20)} & \multicolumn{3}{c}{\textit{Task 4}(40-20)} \\ \cmidrule(lr){2-3} \cmidrule(lr){4-6} \cmidrule(lr){7-9} \cmidrule(lr){10-12}
     & $Acc_r \uparrow$ & $Acc_f \downarrow$ & $Acc_r \uparrow$ & $Acc_f \downarrow$ & $Acc_o \downarrow$ & $Acc_r \uparrow$ & $Acc_f \downarrow$ & $Acc_o \downarrow$ & $Acc_r \uparrow$ & $Acc_f \downarrow$ & $Acc_o \downarrow$ \\ \midrule
    Pre-train &  {63.93} & 66.23 & 64.48 & 57.39 & - & 65.92 & 64.76 & - & 63.80 & 65.16 & - \\
    \midrule
    % \multicolumn{5}{l}{\textcolor{orange}{\textit{(a) continual learning methods}}} \\
    L2$^*$ &  {61.37} & 13.03 &  {66.23} & 11.96 & {33.37}  & {67.62} & 13.71 & {{23.08}} &  {72.75} & 19.47 &{25.73} \\
    EWC$^*$ &  {54.34} & 1.70 &  {61.52} & 4.82 & {2.00}  & {62.85} & 4.80 & {{4.97}} &  {65.04} & 7.63 &{6.65} \\
    MAS$^*$ &  {50.55} & 0.50 &  {54.78} & 0.30 & {0.00}  & {55.29} & 1.00 & {{0.00}} &  {57.54} & 1.80 &{0.40} \\
    LwF &  {58.63} & 1.60 &  {58.49} & 2.41 & {1.50}  & {64.42} & 2.00 & {{1.60}} &  {63.65} & 3.51 &{1.60} \\
    DER &  {46.36} & 15.43 &  {33.59} & 9.65 & {19.84}  & {25.53} & 6.81 & {{12.69}} &  {20.23} & 8.13 &{9.58} \\ 
    DER++ &  {59.68} & 12.73 &  {62.40} & 7.94 & {11.62}  & {66.29} & 7.41 & {{7.98}} &  {68.60} & 12.55 &{7.45} \\ 
    FDR &  {1.60} & 0.30 &  {2.73} & 3.11 & {0.60}  & {6.62} & 2.60 & {{1.00}} &  {12.46} & 0.90 &{1.50} \\
    \midrule
    % \multicolumn{5}{l}{\textcolor{orange}{\textit{(b) machine unlearning methods}}} \\
    Retrain &  {12.31} & 0.00 &  {14.68} & 0.00 & {0.00}  & {17.01} & 0.10 & {{0.00}} &  {19.88} & 1.70 &{0.00} \\
    BAD-T &  43.60 & 0.00 & 45.79 & 0.00 & {0.00}  & 54.01 & 0.00 & {{0.00}} & 53.98 & 0.00 &{0.00} \\
    LIRF$^*$ &  {23.45} & 43.07 & 43.52 & 41.01 & 21.30 & 34.29 & 29.05 & 18.43 &  54.29 & 23.10 & 15.98 \\
    SCRUB &  {46.78} & 0.00 &  {50.48} & 0.00 & {0.00}  & {54.69} & 0.00 & {{0.00}} &  {56.58} & 0.60 &{0.00} \\
    SCRUB-S &  {50.89} & 0.00 &  {52.70} & 0.00 & {0.00}  & {58.40} & 0.00 & {{0.00}} &  {57.68} & 0.00 &{0.00} \\
    % \rowcolor{yellow!50} % 另一行颜色
    GS-LoRA &  {58.25} & 0.00 &  {55.71} & 0.10 & {2.61}  & {0.63} & 0.00 & {{0.00}} &  {2.80} & 0.00 &{0.00} \\ 
    % \rowcolor{yellow!50} %
    GS-LoRA++ &  {58.89} & 0.60 &  {55.53} & 0.20 & {1.90} & {55.49} & 0.00 & {{0.45}} &  {55.23} & 0.00 &{1.40} \\ 
    \midrule
    \rowcolor{gray!30} % 设置当前行颜色
    FG-OrIU  & {60.01} & 0.00  & {56.97} & 0.10 & {0.00} & {60.17} & 0.00 & {{0.00}}  & {61.64} & 0.00 &{0.00} \\
    \bottomrule
    \end{tabular}%
    }
    %}
    \vspace{-2mm}
    \caption{{Incremental Unlearning results for image classification task on OmniBenchmark.} With four tasks, each forgetting 50 classes.}
    \label{tab:cl-recog-caisa100_main_omni}
    % \vspace{-3mm}
\end{table*}

\subsection{Additional Results on Machine Unlearning}
Furthermore, we extend our evaluation to static machine unlearning on CASIA-Face100 and ImageNet-100 by varying the number of forgetting classes across different settings:$\{1,5,10,20,40,60,80,90,95,99\}$. The results for CASIA-Face100 and ImageNet-100 are reported in Tables \ref{tab:single-recog-caisa100_vary_number} and \ref{tab:single-classification_Imagenet-100_vary_number}, respectively.

\begin{table*}[t!]
\centering
\vspace{-5pt}
\resizebox{\textwidth}{!}{
\begin{tabular}{lrrrrrrrrrrrrrrrrr}
\toprule
\multirow{2}{*}{Methods} & \multicolumn{3}{c}{100-1} & \multicolumn{3}{c}{100-5} & \multicolumn{3}{c}{100-10} & \multicolumn{3}{c}{100-20} & \multicolumn{3}{c}{100-40}\\ 
\cmidrule(lr){2-4} \cmidrule(lr){5-7} \cmidrule(lr){8-10} \cmidrule(lr){11-13} \cmidrule(lr){14-16} 
 & $H \uparrow$ & $Acc_r \uparrow$ & $Acc_f \downarrow$ & $H \uparrow$ & $Acc_r \uparrow$ & $Acc_f \downarrow$ & $H \uparrow$ & $Acc_r \uparrow$ & $Acc_f \downarrow$ & $H \uparrow$ & $Acc_r \uparrow$ & $Acc_f \downarrow$ & $H \uparrow$ & $Acc_r \uparrow$ & $Acc_f \downarrow$ \\ \midrule
Pre-train & - & 74.37 & 74.35 & -  & 74.57 & 70.17 & - & 74.44 & 73.67 & - & 73.62 & 74.01 & - & 74.48 & 74.21 \\

% \midrule
\rowcolor{gray!30} % 设置当前行颜色
FG-OrIU & {73.35} & {72.37} & {0.00} & {71.70} & {72.71} & {0.00} & {72.98} & {72.31} & {0.00} & {73.45} & {72.90} & {0.00} & 73.86 & 73.51 & 0.00 \\ 
\midrule
\multirow{2}{*}{Methods} & \multicolumn{3}{c}{100-60} & \multicolumn{3}{c}{100-80} & \multicolumn{3}{c}{100-90} & \multicolumn{3}{c}{100-95} & \multicolumn{3}{c}{100-99}\\ 
\cmidrule(lr){2-4} \cmidrule(lr){5-7} \cmidrule(lr){8-10} \cmidrule(lr){11-13} \cmidrule(lr){14-16} 
 & $H \uparrow$ & $Acc_r \uparrow$ & $Acc_f \downarrow$ & $H \uparrow$ & $Acc_r \uparrow$ & $Acc_f \downarrow$ & $H \uparrow$ & $Acc_r \uparrow$ & $Acc_f \downarrow$ & $H \uparrow$ & $Acc_r \uparrow$ & $Acc_f \downarrow$ & $H \uparrow$ & $Acc_r \uparrow$ & $Acc_f \downarrow$ \\ \midrule
Pre-train & - & 74.77 & 74.11 & -  & 74.47 & 74.32 & - & 74.58 & 74.34 & - & 72.26 & 74.48 & - & 70.48 & 74.41 \\

% \midrule
\rowcolor{gray!30} % 设置当前行颜色
FG-OrIU & {74.03} & {73.95} & {0.00} & {74.54} & {74.76} & {0.00} & {75.72} & {77.15} & {0.00} & {72.66} & {70.92} & {0.00} & 85.33 & 100.00 & 0.00 \\ 

\bottomrule
\end{tabular}%
}
\vspace{-2mm}
\caption{Static Machine Unlearning results for face recognition task on CASIA-Face100 with varying number of forgetting classes.}
\label{tab:single-recog-caisa100_vary_number}
% \vspace{-6mm}
\end{table*}

\begin{table*}[t!]
\centering
\vspace{-5pt}
\resizebox{\textwidth}{!}{
\begin{tabular}{lrrrrrrrrrrrrrrrrr}
\toprule
\multirow{2}{*}{Methods} & \multicolumn{3}{c}{100-1} & \multicolumn{3}{c}{100-5} & \multicolumn{3}{c}{100-10} & \multicolumn{3}{c}{100-20} & \multicolumn{3}{c}{100-40}\\ 
\cmidrule(lr){2-4} \cmidrule(lr){5-7} \cmidrule(lr){8-10} \cmidrule(lr){11-13} \cmidrule(lr){14-16} 
 & $H \uparrow$ & $Acc_r \uparrow$ & $Acc_f \downarrow$ & $H \uparrow$ & $Acc_r \uparrow$ & $Acc_f \downarrow$ & $H \uparrow$ & $Acc_r \uparrow$ & $Acc_f \downarrow$ & $H \uparrow$ & $Acc_r \uparrow$ & $Acc_f \downarrow$ & $H \uparrow$ & $Acc_r \uparrow$ & $Acc_f \downarrow$ \\ \midrule
Pre-train & - & 89.35 &  96.0 & -  & 89.18 & 94.00 & - & 89.47 & 89.00 & - & 88.91 & 90.02 & - & 90.03 & 88.50 \\

% \midrule
\rowcolor{gray!30} % 设置当前行颜色
FG-OrIU & {91.60} & {87.58} & {0.00} & {89.69} & {85.75} & {0.00}  & {88.11} & {87.24} & {0.00} & {88.45} & {86.93} & {0.00} & 86.24 & 84.10 & 0.00 \\ 
\midrule
\multirow{2}{*}{Methods} & \multicolumn{3}{c}{100-60} & \multicolumn{3}{c}{100-80} & \multicolumn{3}{c}{100-90} & \multicolumn{3}{c}{100-95} & \multicolumn{3}{c}{100-99}\\ 
\cmidrule(lr){2-4} \cmidrule(lr){5-7} \cmidrule(lr){8-10} \cmidrule(lr){11-13} \cmidrule(lr){14-16} 
 & $H \uparrow$ & $Acc_r \uparrow$ & $Acc_f \downarrow$ & $H \uparrow$ & $Acc_r \uparrow$ & $Acc_f \downarrow$ & $H \uparrow$ & $Acc_r \uparrow$ & $Acc_f \downarrow$ & $H \uparrow$ & $Acc_r \uparrow$ & $Acc_f \downarrow$ & $H \uparrow$ & $Acc_r \uparrow$ & $Acc_f \downarrow$ \\ \midrule
Pre-train & - & 89.65 & 89.27 & -  & 92.80 & 88.50 & - & 93.00 & 89.00 & - & 94.80 & 89.14 & - & 88.00 & 89.41 \\

% \midrule
\rowcolor{gray!30} % 设置当前行颜色
FG-OrIU & {86.79} & {84.45} & {0.00} & {86.77} & {85.10} & {0.00} & {86.95} & {85.00} & {0.00} & {92.44} & {96.00} & {0.00} & 94.41 & 100.00 & 0.00 \\ 

\bottomrule
\end{tabular}%
}
% \vspace{-2mm}
\caption{Static Machine Unlearning results for image classification task on Imagenet-100 with varying number of forgetting classes.}
\label{tab:single-classification_Imagenet-100_vary_number}
% \vspace{-2mm}
\end{table*}

% \clearpage
% \newpage

\begin{figure*}[!t]
  \centering
   \includegraphics[width=1.0\linewidth]{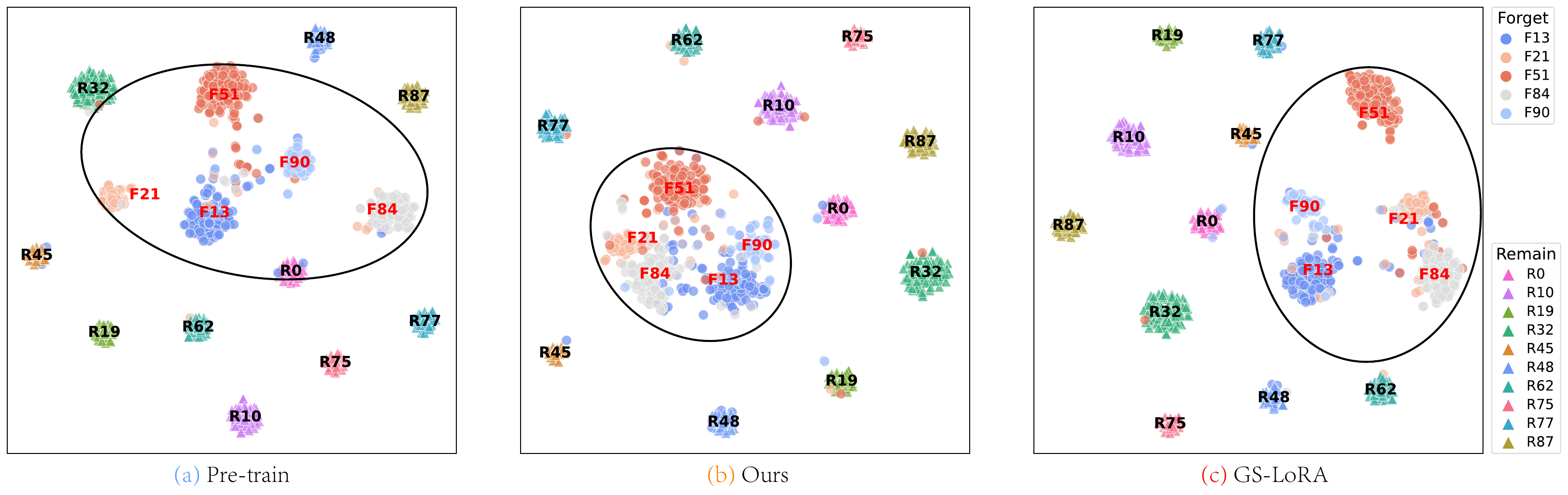}
   % \vspace{-4mm}
   \caption{T-SNE visualizations of the feature distributions for 5 forgetting classes and 10 remaining classes using (a) the Pre-trained model, (b) the unlearned model obtained through our FG-OrIU, and (c) GS-LoRA as feature extractors.}
   \label{apdix:fig:tsne_1}
   \vspace{-2mm}  
\end{figure*}

\begin{figure*}[!t]
  \centering
   \includegraphics[width=1.0\linewidth]{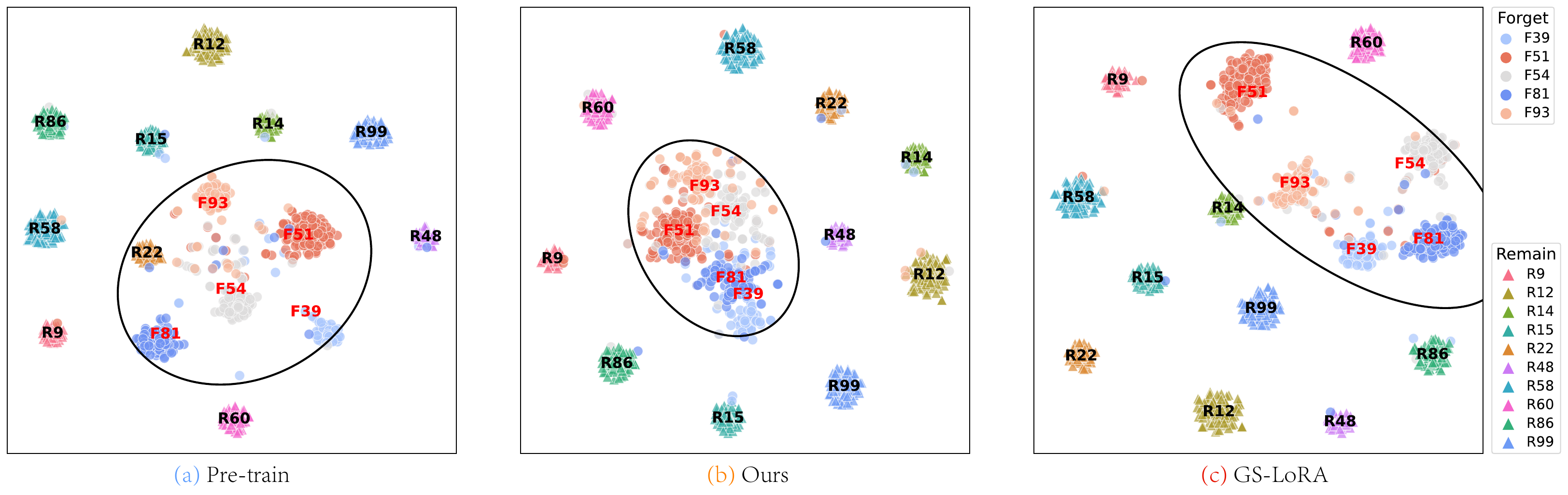}
   \vspace{-4mm}
   \caption{T-SNE visualizations of the feature distributions for 5 forgetting classes and 10 remaining classes using (a) the Pre-trained model, (b) the unlearned model obtained through our FG-OrIU, and (c) GS-LoRA as feature extractors.}
   \label{apdix:fig:tsne_2}
   \vspace{-2mm}  
\end{figure*}

\begin{figure*}[!t]
  \centering
   \includegraphics[width=1.0\linewidth]{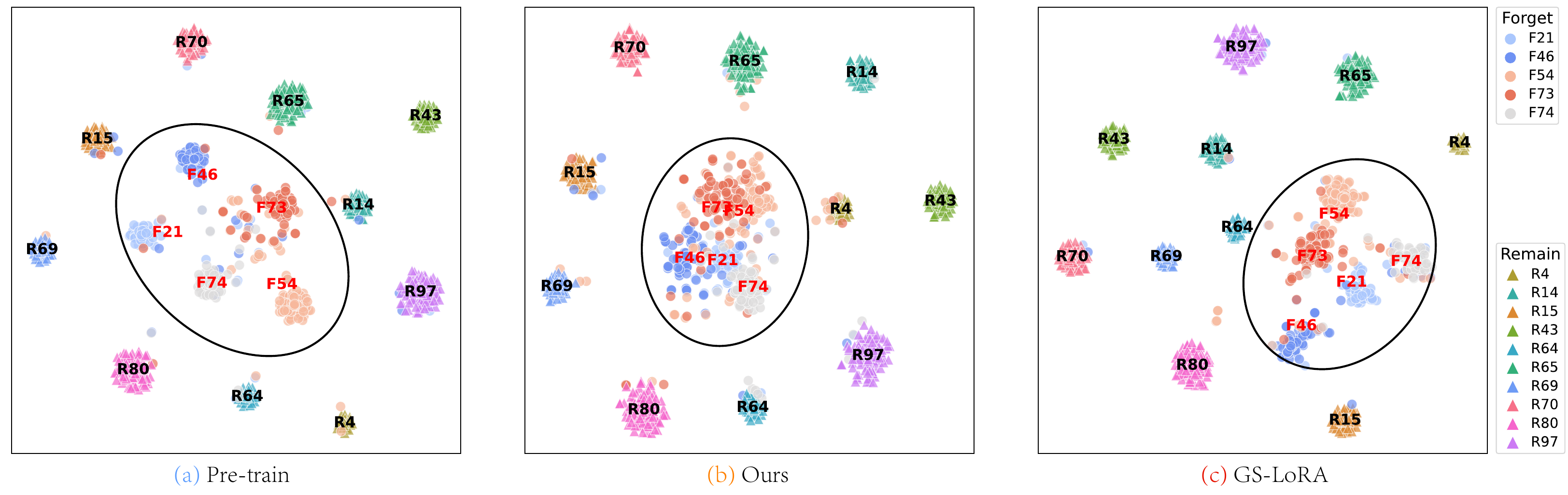}
   \vspace{-4mm}
    \caption{T-SNE visualizations of the feature distributions for 5 forgetting classes and 10 remaining classes using (a) the Pre-trained model, (b) the unlearned model obtained through our FG-OrIU, and (c) GS-LoRA as feature extractors.}
   \label{apdix:fig:tsne_3}
   \vspace{-2mm}  
\end{figure*}

\subsection{More Visualization Results with T-SNE}

\begin{figure*}[t]
  \centering
   \includegraphics[width=0.95\linewidth]{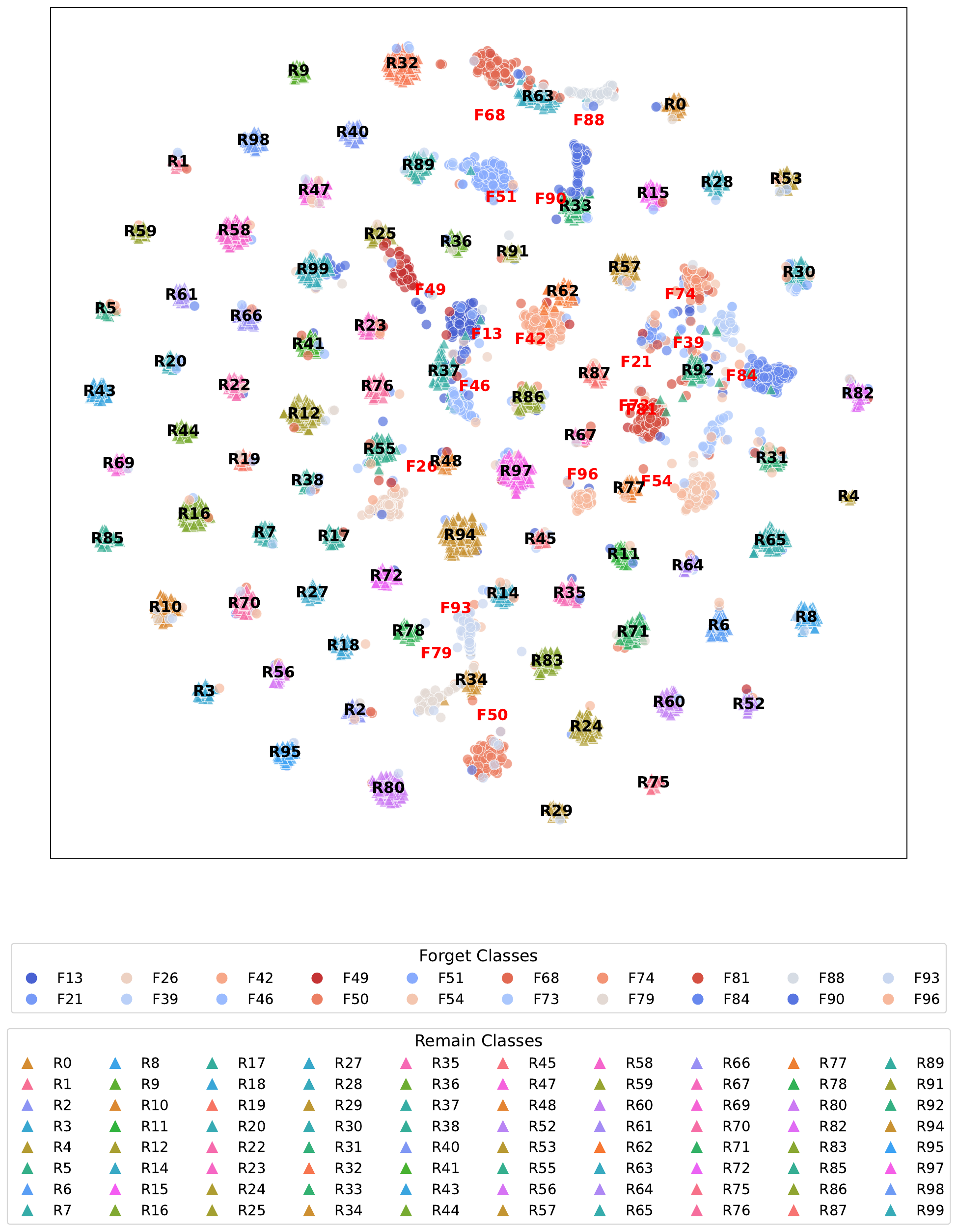}
   % \vspace{-4mm}
    \caption{T-SNE visualizations of the feature distributions for 20 forgetting classes and 80 remaining classes using the unlearned model obtained through GS-LoRA as feature extractors.}
   \label{apdix:fig:tsne_gslora}
   % \vspace{-2mm}  
\end{figure*}

\begin{figure*}[t]
  \centering
   \includegraphics[width=0.95\linewidth]{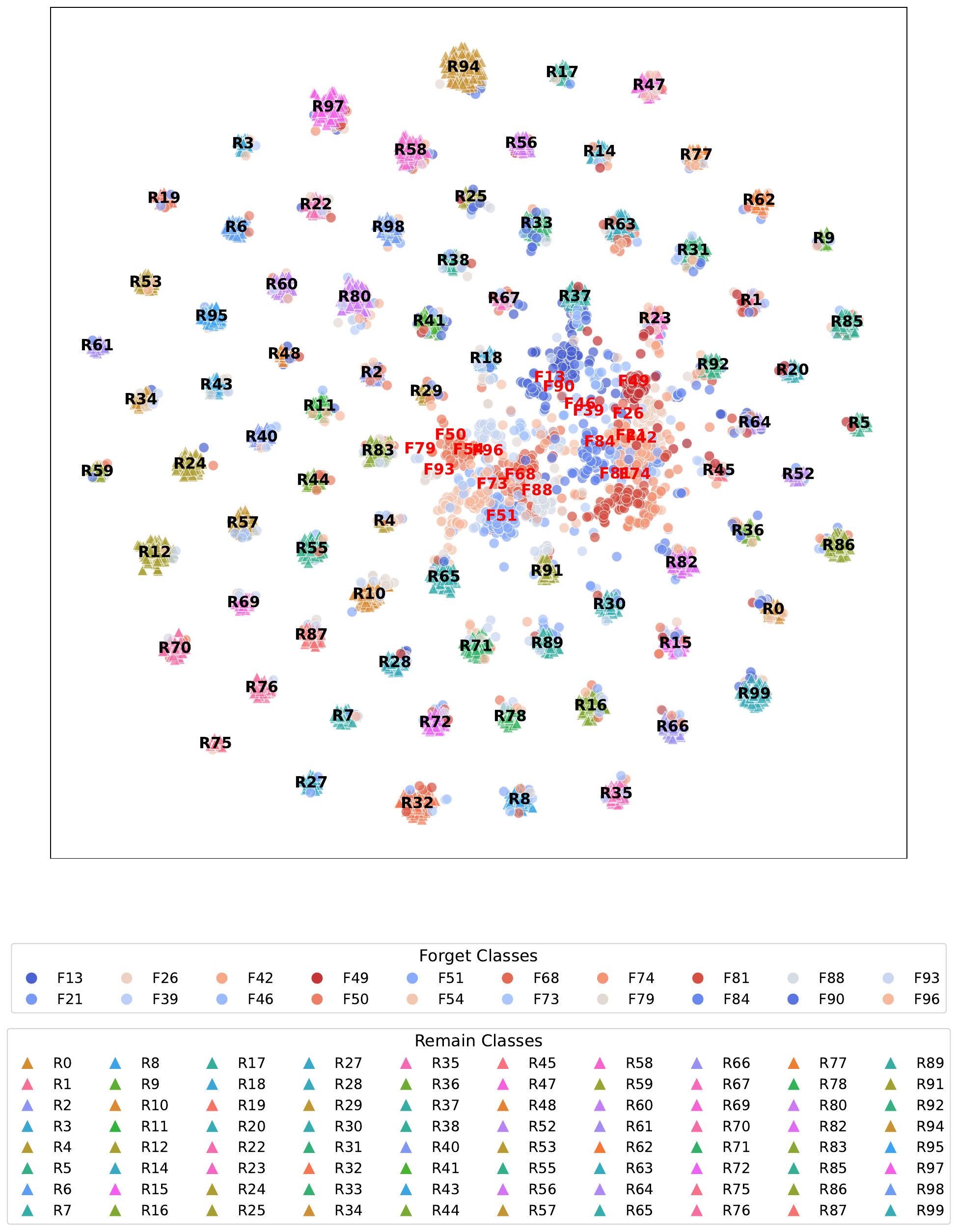}
   % \vspace{-4mm}
    \caption{T-SNE visualizations of the feature distributions for 20 forgetting classes and 80 remaining classes using the unlearned model obtained through our FG-OrIU as feature extractors.}
   \label{apdix:fig:tsne_ours}
   % \vspace{-2mm}  
\end{figure*}

We present visualizations of the feature subspace after unlearning 20 classes from CASIA-Face100. In Figures \ref{apdix:fig:tsne_1}, \ref{apdix:fig:tsne_2}, and \ref{apdix:fig:tsne_3}, we randomly sample five classes from the forgetting set and ten classes from the remaining set to construct a feature matrix, which is then visualized using T-SNE. Figures \ref{apdix:fig:tsne_gslora} and \ref{apdix:fig:tsne_ours} further illustrate the distribution of all 20 forgetting classes and 80 remaining classes when processed by the unlearned models obtained through our FG-OrIU method and GS-LoRA.

Based on these results, we draw the following analysis. Prior works, such as GS-LoRA, impose constraints only at the final output, which results in the unlearned model extracting degraded features for the forgetting classes. These degraded features become easily entangled with those from the remaining classes, leading to a dilemma where the method either fails to completely forget or significantly degrades the performance on the remaining classes. In contrast, our FG-OrIU method directly enforces constraints at both the feature and gradient levels for both forgetting and remaining classes, achieving thorough unlearning while effectively preventing feature entanglement between the two sets. Furthermore, Figures \ref{apdix:fig:tsne_1}, \ref{apdix:fig:tsne_2}, and \ref{apdix:fig:tsne_3}, reveal that compared to the pre-trained model, our method enforces constraints on features in a way that results in a more compact decision boundary. This compactness reduces misclassifications, leading to slight improvements in certain cases over the original pre-trained model.

% \clearpage
\newpage
\section{Algorithm}
\label{append:algho}
\begin{algorithm}[!t]
\caption{Pseudo code of Feature Subspace Decomposition and Dynamic Subspace Adaptation}
\label{alg:FG-OrIU}
\begin{algorithmic}
\STATE \textbf{Input:} Pre-trained model $\mathcal{M}$ and its layers $N$, the Pre-trained model's function $f_{\mathcal{M}}$, LoRA module $\bm{\theta}$, a sequence of unlearning task $\{\mathcal{T}_1,\mathcal{T}_2,\cdots,\mathcal{T}_{T}\}$, forgetting classes $\mathbf{D}_{f,t}$ and  remaining classes $\mathbf{D}_{r,t}$ for each unlearning task, threshold $\epsilon$, hyper-parameter $\lambda_1$ and $\lambda_2$, learning rate $\eta$

% \STATE Collect augmentation sentences $\{s_i\}_{i=0}^{N}$ and corresponding token indices $id_{i,j}$ related to target concept $\mathcal{X}_t$
\FOR{$t$ in $\{1,\dots T\}$}

\IF {$t=1$}
    \STATE \textbf{Stage One: Feature Subspace Decomposition}
    \FOR{$l$ in $\{1,\dots,N\}$}
        \STATE \textit{build forgetting feature subspace}
        \STATE $\bm{R}^{l}_{f,1} \gets$ $f_{\mathcal{M}}(\mathbf{D}_{f,1})$
        \STATE $\bm{U}_{f,i}^{l}\bm{\Sigma}_{f,i}^{l}(\bm{V}_{f,i}^{l})^{T} \gets \text{SVD}(\bm{R}^{l}_{f,1})$
        \STATE $\bm{B}^{l}_{f,1}  \gets \bm{U}_{f,i}^{l}[:k]$
        \STATE $\mathcal{S}^{l}_{f,1}=\text{span}\{\bm{B}^{l}_{f,1}\}$
        \STATE \textit{build remaining feature subspace}
        \STATE $\bm{R}^{l}_{r,1} \gets$ $f_{\mathcal{M}}(\mathbf{D}_{r,1})$
        \STATE $\bm{U}_{r,i}^{l}\bm{\Sigma}_{r,i}^{l}(\bm{V}_{r,i}^{l})^{T} \gets \text{SVD}(\bm{R}^{l}_{r,1})$
        \STATE $\bm{B}^{l}_{r,1}  \gets \bm{U}_{r,i}^{l}[:k]$
        \STATE $\mathcal{S}^{l}_{r,1}=\text{span}\{\bm{B}^{l}_{r,1}\}$
    \ENDFOR
\ELSE
    \STATE \textbf{Stage Three: Dynamic Subspace Adaptation}
        \STATE \textit{update forgetting feature subspace}
        % \STATE $\bm{R}^{l}_{f,i} \gets f_{\mathcal{M}}(\mathbf{D}_{f,i})$
        % \STATE ${\hat{\bm{R}}}^{l}_{f,i+1} \gets \bm{R}^{l}_{f,i+1} - \bm{B}^{l}_{f,i} \left(\bm{B}^{l}_{f,i}\right)^{T}  \left(\bm{R}^{l}_{f,i+1}\right)$
        % \STATE $\hat{\bm{U}}^{l}_{f,i+1}\hat{\bm{\Sigma}}^{l}_{f,i+1}(\hat{\bm{V}}^{l}_{f,i+1})^{T} \gets \text{SVD}({\hat{\bm{R}}}^{l}_{f,i+1})$
        % \STATE $ \{\bm{u}_{1}, \dots, \bm{u}_{h}\} \gets \hat{\bm{U}}^{l}_{f,i+1}[:h]$
        \STATE $\mathcal{S}^{l}_{f,i} \gets$ via Eq.(\ref{subspace_update})
        \STATE \textit{update remaining feature subspace}
        % \STATE $\bm{R}^{l}_{r,i} \gets f_{\mathcal{M}}(\mathbf{D}_{r,i})$
        % \STATE ${\hat{\bm{R}}}^{l}_{r,i+1} \gets \bm{R}^{l}_{r,i+1} - \bm{B}^{l}_{r,i} \left(\bm{B}^{l}_{r,i}\right)^{T}  \left(\bm{R}^{l}_{r,i+1}\right)$
        % \STATE $\hat{\bm{U}}^{l}_{r,i+1}\hat{\bm{\Sigma}}^{l}_{r,i+1}(\hat{\bm{V}}^{l}_{r,i+1})^{T} \gets \text{SVD}({\hat{\bm{R}}}^{l}_{r,i+1})$
        % \STATE $ \{\bm{u}_{1}, \dots, \bm{u}_{h}\} \gets \hat{\bm{U}}^{l}_{r,i+1}[:h]$
        \STATE $\mathcal{S}^{l}_{r,i} \gets$ via Eq.(\ref{subspace_update})
\ENDIF
\STATE \textbf{Stage Two: Dual Orthogonal Projection}
\FOR{each batch $\left(\bm{x}_f,\bm{x}_r\right)$ in $\left(\mathbf{D}_{f,t}, \mathbf{D}_{r,t}\right)$}
    \STATE \textit{feature orthogonal projection}
    \STATE $\mathcal{L}_{fop-f} \gets $ via Eq.(\ref{loss:L_fop-f})
    \STATE $\mathcal{L}_{fop-r} \gets $ via Eq.(\ref{loss:L_fop-r})
    \STATE $\mathcal{L}_{ce} \gets $ via Eq.(\ref{loss:L_ce})
    \STATE $\mathcal{L}_{total} \gets$ via Eq.(\ref{eq:total_loss})
    \FOR{$l$ in $\{1,\dots,N\}$}
        \STATE \textit{gradient orthogonal projection}
        \STATE $\bm{g_{l}} \gets \nabla_{\theta_{l}}\mathcal{L}_{total}$ 
        \STATE $\bm{\hat{g}_{l}} \gets $ via Eq.(\ref{eq:original approx unlearn steepest descent})
        \STATE $\bm{\theta}_{l} \gets \bm{\theta}_{l} -\eta \bm{\hat{g}_{l}}$
    \ENDFOR
\ENDFOR
\ENDFOR
\end{algorithmic}
\end{algorithm}

\section{Details about DIP} 
\label{app:dip}
% In this section，我们首先this paper 使用的用于将特征重建为图像的方法（后面简称为图像重建方法）, \ie, 详细介绍其背景，以及其能被用于揭示 feature中的语义信息的原理 in Section \ref{app:dip_intro}. We then 详细地介绍了我们的实现方式以及对 reconstructed image进行评估的方法, from 定性和定量，以及介绍了所使用的定量指标SSIM和PSNR in Section

\subsection{DIP and Image Reconstruction}
\label{app:dip_intro}
Deep Image Prior (DIP) \cite{ulyanov2018deep} leverages the inherent structure of convolutional neural networks (CNNs) as an implicit prior for natural images by optimizing a randomly initialized CNN to reconstruct a target image from noise, progressively capturing structured information before overfitting to noise. This self-supervised approach exploits the inductive bias of CNNs without requiring external datasets or supervision, making it effective for image restoration tasks such as denoising, inpainting, and super-resolution \cite{gandelsman2019double,ren2020neural,mataev2019deepred,fermanian2021regularizing}. By aligning reconstructed images with feature representations from pre-trained models, DIP provides insights into the semantic information encoded in the extracted features.

In our study, we follow \cite{zhao2020makes, yoo2023improving} to investigate the semantic information encoded in the features extracted from the last layers of the pre-trained model and the unlearned model. Specifically, we reconstruct images from features corresponding to forgetting classes in the unlearned model using DIP, then evaluate whether these reconstructions retain semantic attributes of the forgetting classes through (1) qualitative visual analysis (e.g., visualization of the reconstructed image) and (2) quantitative assessment using SSIM and PSNR metrics (the definition is as follows). These approaches enable systematic evaluation of whether features in unlearned models retain traces of semantic information from intentionally forgotten classes.

\noindent
\textbf{SSIM (Structural Similarity Index)}

The Structural Similarity Index (SSIM) measures the similarity between two images. It is defined as follows:
\begin{equation}
SSIM(x, y) = \frac{(2\mu_x \mu_y + C_1)(2\sigma_{xy} + C_2)}
{(\mu_x^2 + \mu_y^2 + C_1)(\sigma_x^2 + \sigma_y^2 + C_2)}
\end{equation}
where \( x \) and \( y \) are two images (or image patches) being compared. \( \mu_x \) and \( \mu_y \) are the mean intensities of \( x \) and \( y \), respectively:
\begin{equation}
\mu_x = \frac{1}{N} \sum_{i=1}^{N} x_i, \quad
\mu_y = \frac{1}{N} \sum_{i=1}^{N} y_i
\end{equation}
where \( \sigma_x^2 \) and \( \sigma_y^2 \) are the variances of \( x \) and \( y \):
\begin{equation}
\sigma_x^2 = \frac{1}{N-1} \sum_{i=1}^{N} (x_i - \mu_x)^2, \quad
\sigma_y^2 = \frac{1}{N-1} \sum_{i=1}^{N} (y_i - \mu_y)^2
\end{equation}
where \( \sigma_{xy} \) is the covariance between \( x \) and \( y \):
\begin{equation}
\sigma_{xy} = \frac{1}{N-1} \sum_{i=1}^{N} (x_i - \mu_x)(y_i - \mu_y)
\end{equation}
where \( C_1 \) and \( C_2 \) are small constants to stabilize the division, typically defined as:
\begin{equation}
C_1 = (K_1 L)^2, \quad C_2 = (K_2 L)^2
\end{equation}
where \( L \) is the dynamic range of pixel values (e.g., 255 for an 8-bit image), and \( K_1 \) and \( K_2 \) are small constants, commonly set to 0.01 and 0.03, respectively.

\noindent
\textbf{PSNR (Peak Signal-to-Noise Ratio)}

The Peak Signal-to-Noise Ratio (PSNR) is a metric used to measure the quality of an image compared to a reference. It is defined as:

\begin{equation}
PSNR = 10 \log_{10} \left( \frac{L^2}{MSE} \right)
\end{equation}
where: MSE (Mean Squared Error) is calculated as:
\begin{equation}
MSE = \frac{1}{MN} \sum_{i=1}^{M} \sum_{j=1}^{N} (x_{i,j} - y_{i,j})^2
\end{equation}
where \( x_{i,j} \) and \( y_{i,j} \) are the pixel values of the reference and distorted images, respectively. And \( L \) is the maximum possible pixel value (e.g., 255 for an 8-bit image).

\subsection{Implementation of DIP}

We follow \cite{zhao2020makes, yoo2023improving} and the open-source implementation \footnote{https://github.com/atiyo/deep\_image\_prior} to implement DIP. Specifically, we use an autoencoder as the trainable network, with an encoder composed of four convolutional layers with increasing channel depths (64, 128, 256, and 512), each followed by a ReLU activation, and a decoder symmetric to the encoder structure but utilizing transposed convolutional layers. The final layer applies a Sigmoid activation to constrain pixel values between 0 and 1, ensuring a proper reconstructed image output.

\begin{figure*}[t]
  \centering
   \includegraphics[width=1\linewidth]{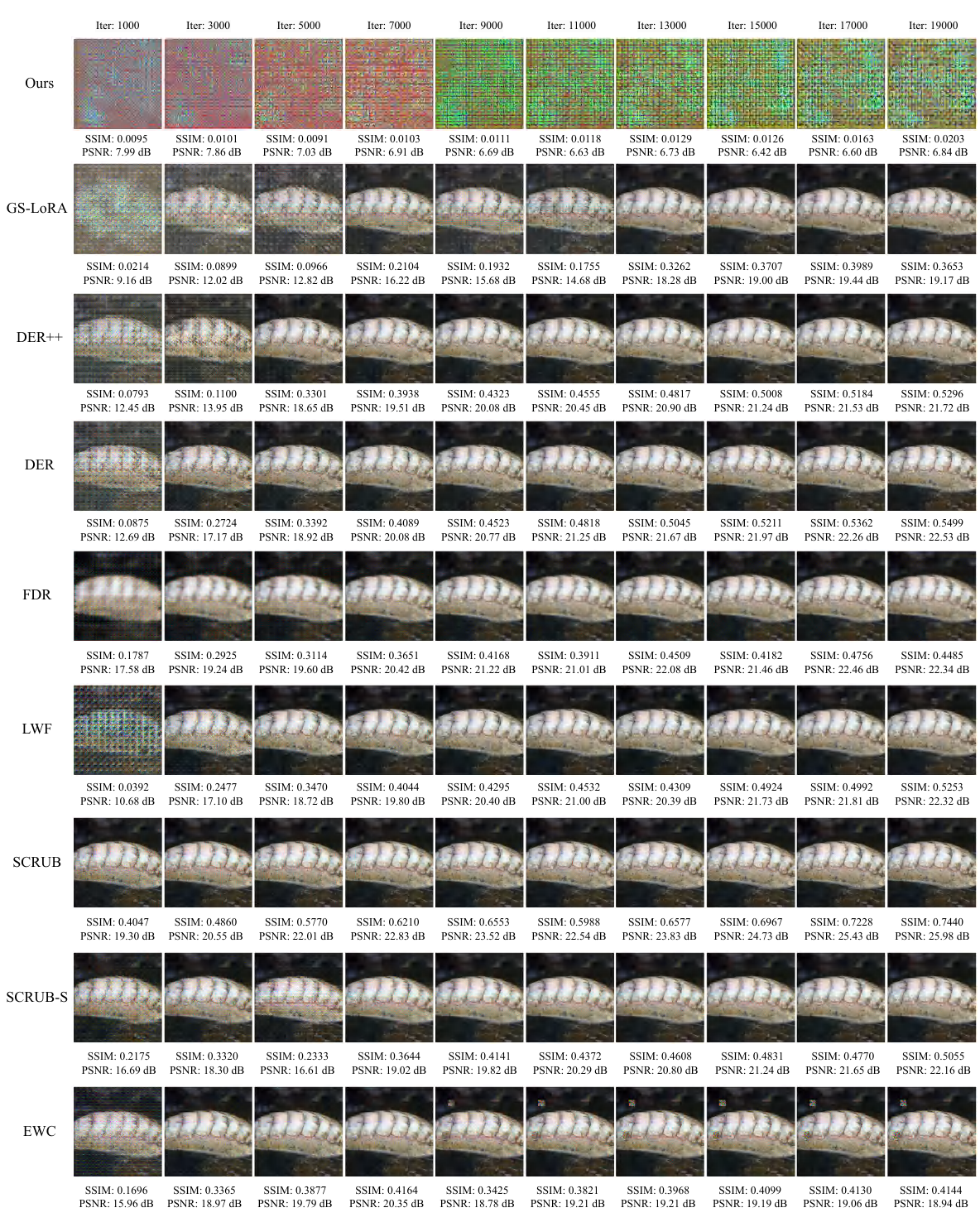}
   \vspace{-4mm}
\caption{The image reconstruction process of example 1 from forgetting classes.}
 \label{apdix:fig:dip_process_exp1}
   \vspace{-2mm}  
\end{figure*}

\begin{figure*}[t]
  \centering
   \includegraphics[width=1\linewidth]{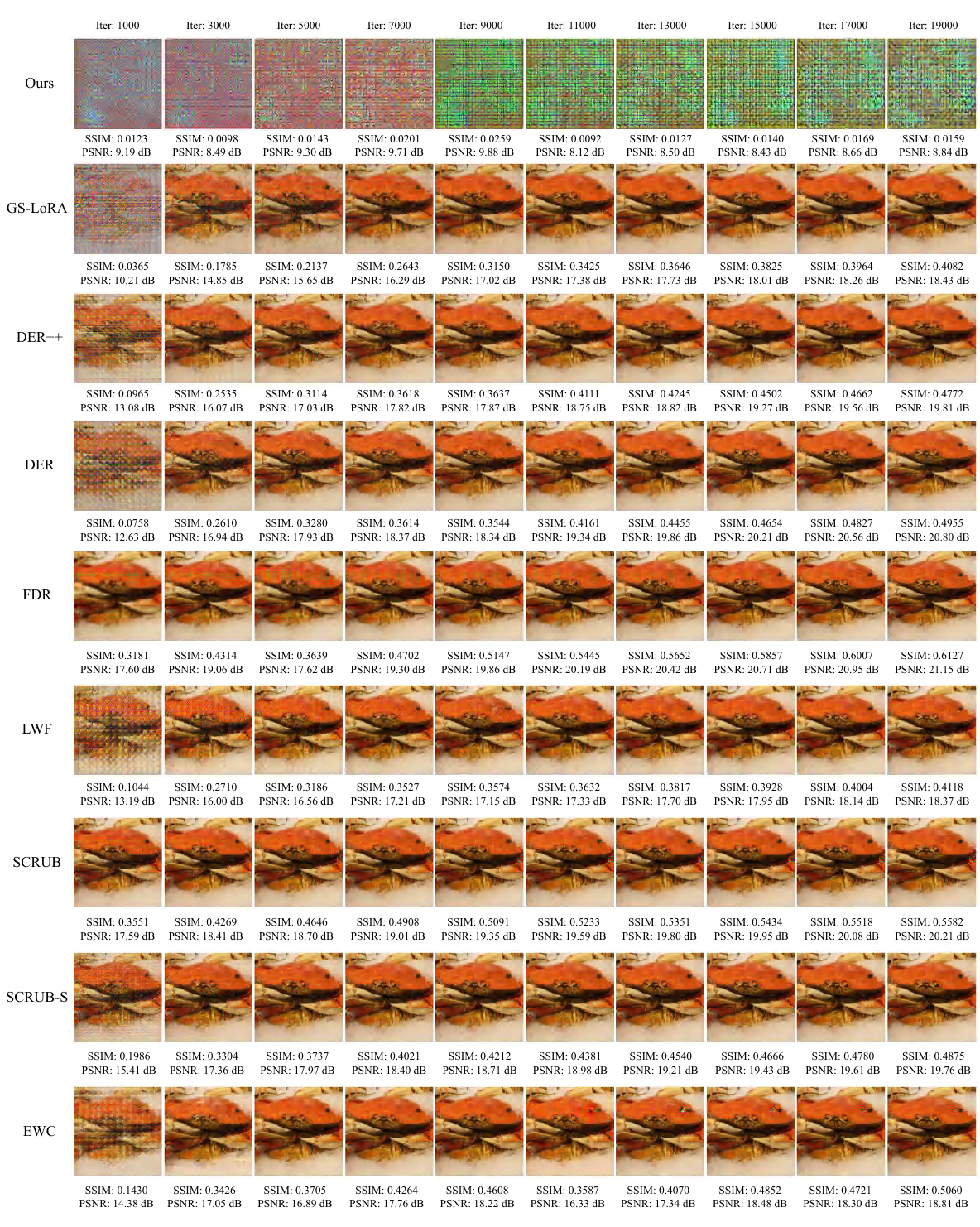}
   \vspace{-4mm}
    \caption{The image reconstruction process of example 2 from forgetting classes.}
   \label{apdix:fig:dip_process_exp2}
   \vspace{-2mm}  
\end{figure*}

During training, we use the unlearned model $f_{\mathcal{M}}$ to extract the feature from a sample belonging to the forgetting class as the training target, $f_{\mathcal{M}}(\bm{x}_f)$. We initialize a fixed noise $\bm{x}_{noise}$ with the same size as the sample (or resize it accordingly) and feed the noise into the AE network to obtain the output $f_{AE}(\bm{x}_{noise})$. We then extract features from $f_{AE}(\bm{x}_{noise})$ using the unlearned model $f_{\mathcal{M}}(\bm{x}_f)$ to obtain $f_{\mathcal{M}}(f_{AE}(\bm{x}_{noise}))$. Finally, the AE network is updated by minimizing the following loss function:
\begin{align} 
\mathcal{L}_{recon} = ||f_{\mathcal{M}}(f_{AE}(\bm{x}_{noise})) - f_{\mathcal{M}}(\bm{x}_f)||_{F}^{2} \end{align}
The reconstructed image is output every fixed number of iterations. We obtain the unlearned model using different IU methods and reconstruct two examples, with the reconstruction process illustrated in Figure \ref{apdix:fig:dip_process_exp1} and Figure \ref{apdix:fig:dip_process_exp2}, respectively. In addition, we provide SSIM and PSNR metrics during the reconstruction process to quantitatively evaluate the reconstructed images.

\subsection{More Visualization Results}

\begin{figure*}[t]
  \centering
   \includegraphics[width=1\linewidth]{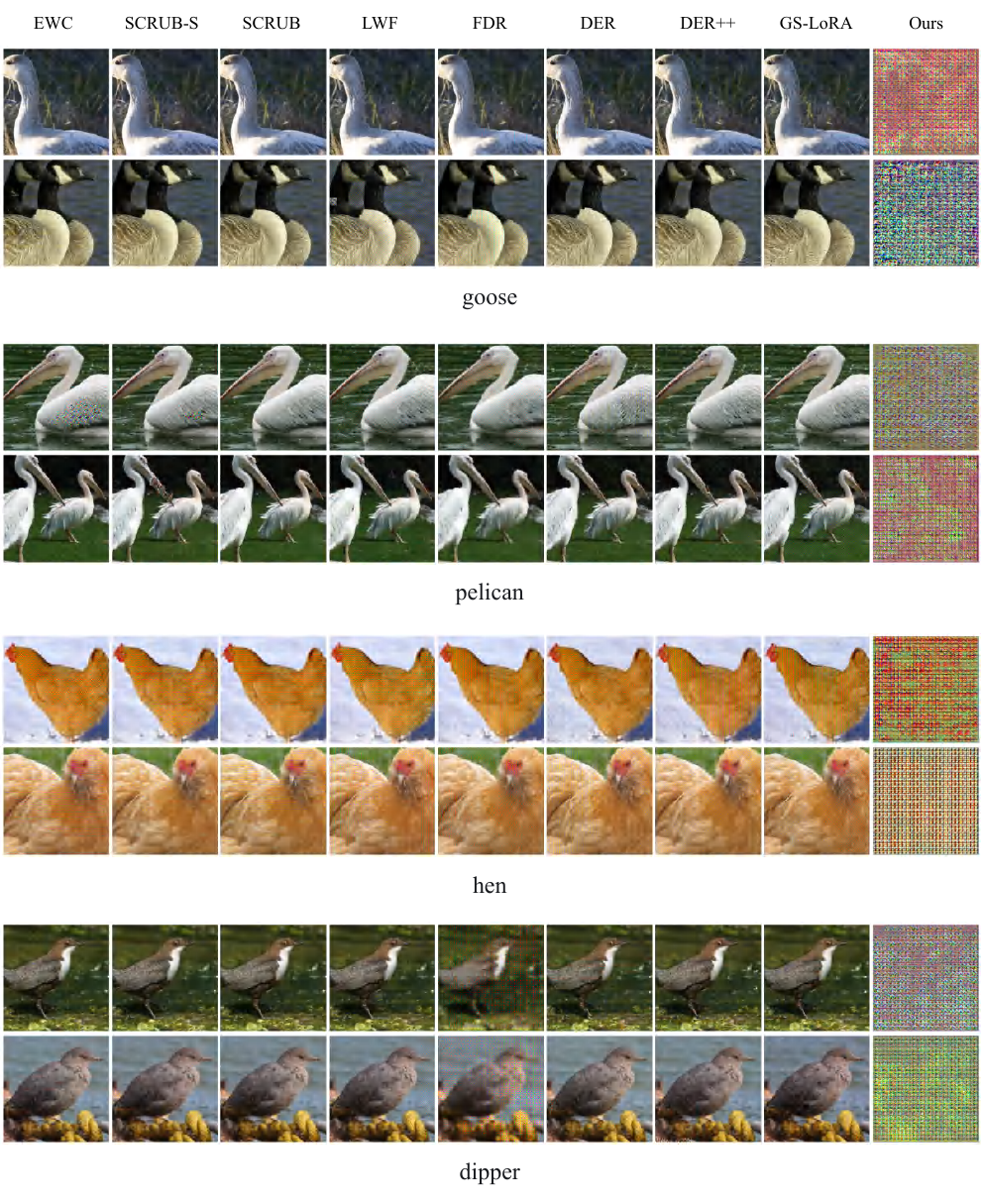}
   \vspace{-4mm}
    \caption{The image reconstruction of forgetting classes' samples.}
   \label{apdix:fig:dip_forget}
   \vspace{-2mm}  
\end{figure*}

\begin{figure*}[t]
  \centering
   \includegraphics[width=1\linewidth]{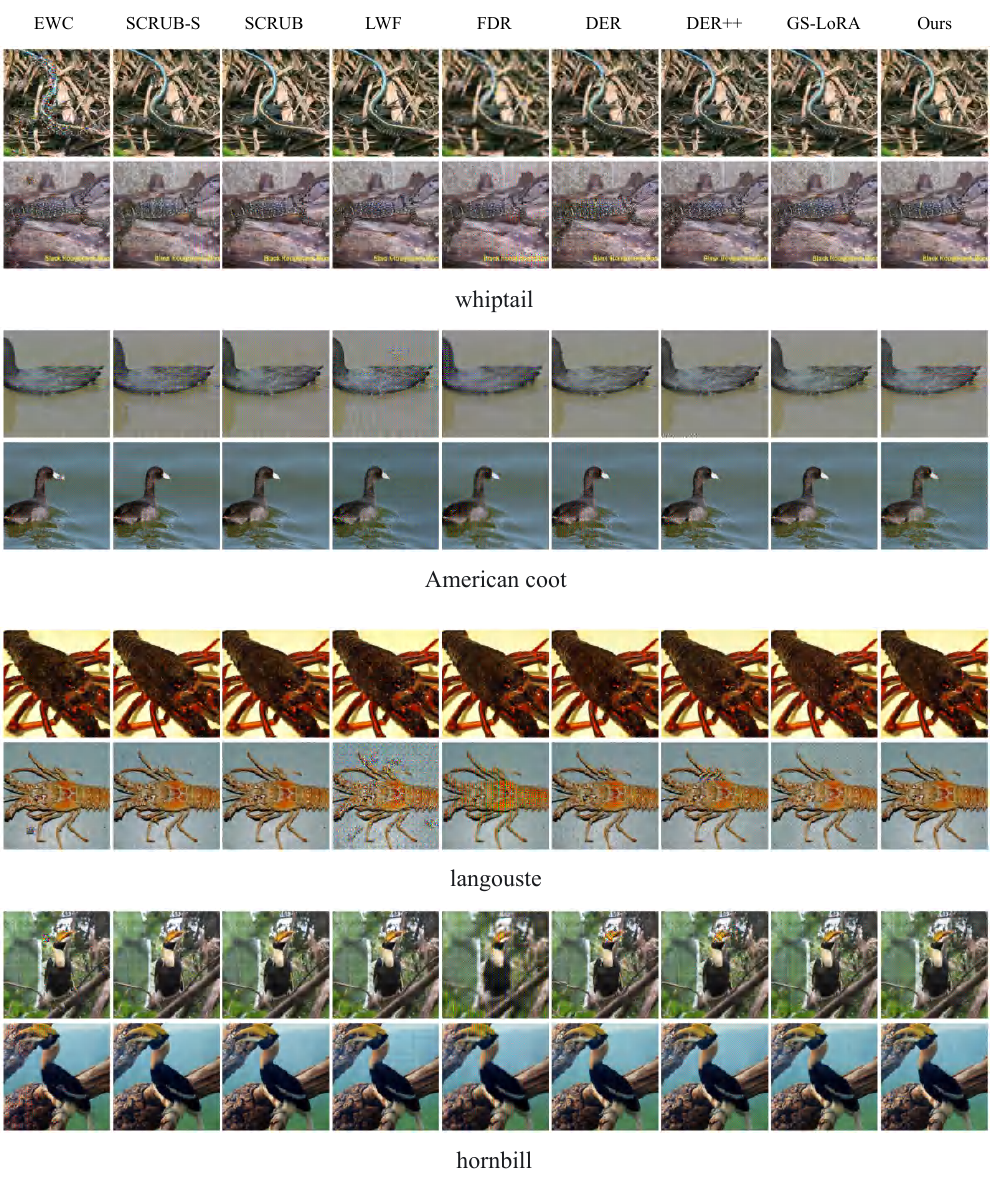}
   \vspace{-4mm}
    \caption{The image reconstruction of remaining classes' samples.}
    \label{apdix:fig:dip_remain}
   \vspace{-2mm}  
\end{figure*}

Here, we provide additional visualization results of DIP on both samples from forgetting classes and remaining classes. We select two samples from each of the four forgetting classes and two samples from each of the four remaining classes. The DIP reconstruction results are presented in Figure \ref{apdix:fig:dip_forget} and Figure \ref{apdix:fig:dip_remain}, respectively.

\end{document}